\documentclass[final]{cvpr}

\usepackage{times}
\usepackage{epsfig}
\usepackage{graphicx}
\usepackage{amsmath}
\usepackage{amssymb}

\providetoggle{cvprfinal}
\settoggle{cvprfinal}{true}

\usepackage[pagebackref=true,breaklinks=true,colorlinks,bookmarks=false,hypertexnames=false]{hyperref}
\iftoggle{cvprfinal}{
\renewcommand*{\backref}[1]{\ifx#1\relax \else Page #1 \fi}
\renewcommand*{\backrefalt}[4]{%
    \ifcase #1 \footnotesize{(Not cited.)}%
    \or        \footnotesize{(Cited on pg~#2.)}%
    \else      \footnotesize{(Cited on pg~#2.)}%
    \fi}
}{}

\def\cvprPaperID{5690} 
\def\confYear{CVPR 2021}

\usepackage{float}

\usepackage[dvipsnames]{xcolor}

\usepackage[capitalise]{cleveref}
\usepackage{placeins} 

\renewcommand*{\backrefalt}[4]{%
    \ifcase #1 \footnotesize{}%
    \or        \footnotesize{(Cited on page~#2.)}%
    \else      \footnotesize{(Cited on pages~#2.)}%
    \fi}
\usepackage{changepage}

\newcommand{\R}{\mathbb{R}}
\providecommand{\argmin}{\mathop\mathrm{arg min}}

\usepackage{etoc}

\usepackage[numbers,sort&compress]{natbib}
\usepackage{booktabs}

\newcommand{\resetcounters}[0]{
    \setcounter{table}{0}
    \setcounter{figure}{0}
    \renewcommand{\thefigure}{\thesection\arabic{figure}}
}
\Crefname{section}{Sec}{Secs.}
\Crefname{figure}{Fig}{Figs.}
\usepackage{fontawesome}

\begin{document}


\title{Matched sample selection with GANs for mitigating attribute confounding}

\author{Chandan Singh$^{1, 4}$ \and Guha Balakrishnan$^{2, 4}$ \and Pietro Perona$^{3, 4}$ \\

\and 
$^1$ University of California at Berkeley\\
$^2$ Massachusetts Institute of Technology\\
$^3$ California Institute of Technology \\ 
$^4$ Amazon Web Services \\}

\maketitle

\begin{abstract}
    Measuring biases of vision systems with respect to protected attributes like gender and age is critical as these systems gain widespread use in society. However, significant correlations between attributes in benchmark datasets make it difficult to separate algorithmic bias from dataset bias. To mitigate such attribute confounding during bias analysis, we propose a \textbf{matching}~\cite{rubin1973matching} approach that selects a subset of images from the full dataset with balanced attribute distributions across protected attributes. Our matching approach first projects real images onto a generative adversarial network (GAN)'s latent space in a manner that preserves semantic attributes. It then finds image matches in this latent space across a chosen protected attribute, yielding a dataset where semantic and perceptual attributes are balanced across the protected attribute.
    
    We validate projection and matching strategies with qualitative, quantitative, and human annotation experiments. We demonstrate our work in the context of gender bias in multiple open-source facial-recognition classifiers and find that bias persists after removing key confounders via matching.\footnote{Code and documentation to reproduce the results here and apply the methods to new data is available \iftoggle{cvprfinal}{at \href{https://github.com/csinva/matching-with-gans}{\faGithub~github.com/csinva/matching-with-gans}.}{on github (link anonymized, see zip file).}
    We additionally release all collected annotations and computed intermediate outputs.} 
\end{abstract}

\etocdepthtag.toc{mtchapter}
\etocsettagdepth{mtchapter}{subsection}
\etocsettagdepth{mtappendix}{none}

\section{Introduction}


Computer vision systems have applications in the entertainment, education, consumer, medical, security, and policing fields. In many applications, these systems can be an important factor used by humans to make impactful decisions. It is therefore important to minimize their potential biases with respect to protected attributes such as sex, gender, national origin, ethnicity, and age. 

\begin{figure}[t]
    \centering
    \includegraphics[width=\columnwidth]{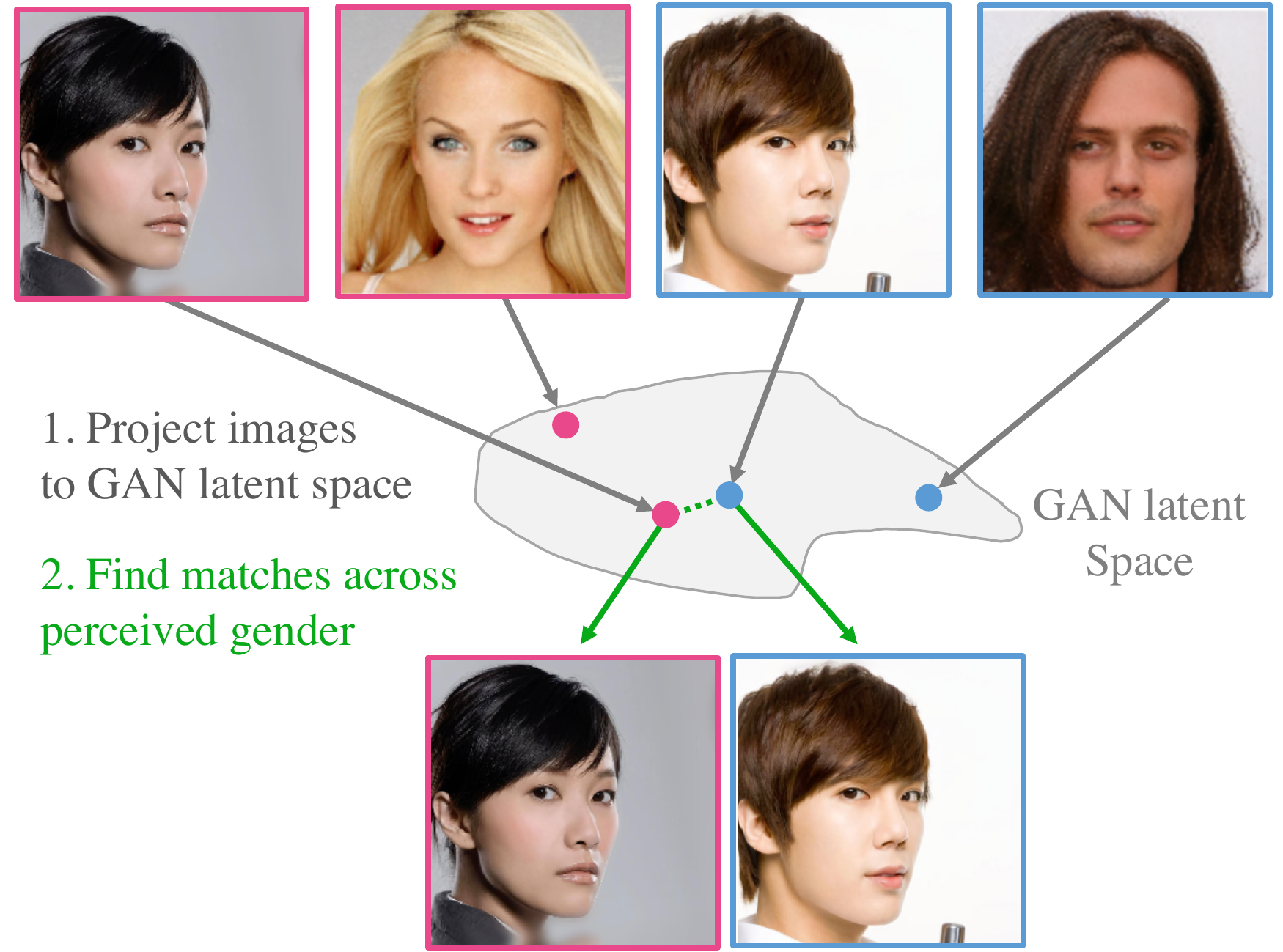}
    \caption{\textbf{Finding matched samples by GAN latent-space projection.} Given face images we look for a `matched sample', i.e. a pair of face images that has similar attributes apart from one attribute (perceived gender here) which needs to be different. We propose a method for matching in GAN latent space (\cref{sec:matching}) which requires projecting real face images to that space (\cref{sec:projection}). The resulting matched samples can be used for benchmarking bias with respect to the protected attribute.}
    \label{fig:overview}
\end{figure}

A key step in minimizing bias is measuring it. Benchmarking bias is not straightforward due to the presence of confounders in datasets. For example, male celebrities in the CelebA-HQ dataset~\cite{karras2017progressive, liu2015faceattributes} are on average older and darker-skinned than their female counterparts; lighting, pose, expression and background may also be systematically different. These confounding factors can spuriously allocate inaccuracies in the model to the wrong attribute. If one finds a higher recognition error for, say, younger people, it is difficult to tell whether it is age or gender (females being on average younger) that causes the bias~\cite{balakrishnan2020causal}. Pinpointing the cause of the bias serves as the starting point of a process to eliminate the bias, by applying appropriate corrective steps during algorithm training and/or collecting additional training data with appropriate statistics.


Random sampling of test data `in the wild' is well-known to produce test sets that are not appropriate for causal inference due to spurious uncontrolled correlations as explained above.
A central concept in causal inference, used in clinical trials as well as experiments across domains is \textit{causal matching}~\cite{rubin1973matching,stuart2010matching}, which selects unconfounded image pairs (i.e. the pairs are similar in aspects besides a specified protected attribute; see \cref{fig:overview}).These pairs are then aggregated to yield a subset of the original dataset where semantic attributes are balanced across groups, thereby mitigating potential confounding. However, the matching process is difficult because, unlike with low-dimensional tabular data, image features are difficult and too numerous to specify and explicitly annotate.

Our approach addresses this challenge by exploiting the properties of the latent space of a pretrained GAN in a domain of interest, e.g., faces. Recent studies have demonstrated not only that GANs can produce remarkably photorealistic and diverse image samples, but that their latent spaces disentangle semantic dimensions of the domain ~\cite{goodfellow2014generative,karras2017progressive,karras2019style,karras2019analyzing}. Our method consists of two components: projecting images onto a GAN's latent space, and finding close matches in the latent space.

We improve upon an existing technique for projecting real images onto StyleGAN2's latent ``style'' space~\cite{abdal2019image2stylegan, karras2019analyzing} by introducing a regularization penalty. We then show how to perform two matching strategies using this latent space: GAN-distance matching and propensity-score matching. We demonstrate with experiments that both approaches can balance protected attributes such as race and age across subgroups, despite having no explicit access to these attribute labels. We then briefly show how the balanced data can be used to measure algorithmic bias -- we benchmark bias of several open-source facial recognition systems with respect to a person's perceived gender\footnote{We use the gender labels that were manually annotated in the curation of the CelebA dataset~\cite{liu2015faceattributes}. These annotations do not necessarily reflect the {\em gender identity} of the person, rather they refer to {\em binarized gender as perceived by a casual observer}. While this difference is immaterial to the techniques we propose, we prefer to make this transparent and therefore we refer to `perceived gender' rather than `gender'.}. We find that these systems are worse at identifying faces perceived as female, solidifying evidence measured on the original, unmatched data (\cref{fig:bias_benchmarking}). 




This work's main contribution is a GAN-based method for producing a dataset of matched samples culled from a larger dataset of {\em real} face images (i.e., not synthetic). Each pair of samples differs by one selected attribute (e.g., gender) and is as similar as possible with respect to all other attributes (\cref{sec:matching}). Additional contributions include (a) A method for projecting real images onto the latent space of StyleGAN2 (\cref{sec:projection}), evaluated with detailed human experiments. (b) Experimental evaluation of our matching strategy showing that it successfully balances key covariates. (c) an application to measuring algorithmic bias in face recognition  (\cref{sec:benchmarking}) where we detect gender-based bias in academic-grade algorithms. 


\section{Related work}

\paragraph{Matching and causal inference}

Matching is a popular and well-established technique in the field of causal inference~\cite{stuart2010matching}; it is used broadly across a variety of fields including statistics~\cite{rosenbaum2002overt,rubin2006matched}, epidemiology~\cite{brookhart2006variable}, sociology~\cite{morgan2006matching}, economics~\cite{imbens2004nonparametric}, and political
science~\cite{ho2007matching}. Our work builds on these ideas, but measures distance for matching using a pretrained neural network (StyleGAN2). This choice is motivated by observations that the StyleGAN2 latent space captures a wide variety of useful semantic attributes without any supervision.

Alternative causal analyses may require explicit values of the features over which we are trying to balance, such as age, background, and pose, which may be extremely difficult to collect and accurately measure. Moreover, most approaches, such as propensity score matching~\cite{dehejia2002propensity}, do not work well when applied directly to low-level features like raw pixels. Classifying the probability of membership to a group, e.g., images perceived as male, directly from images is a difficult task for which models have shown considerable bias. Our approach overcomes this challenge by first embedding images in the StyleGAN2 latent space, in which a simple linear classifier can easily predict group membership.

One recent work has proposed using neural networks to aid in causal matching~\cite{ramachandra2018deep} on low-dimensional, toy examples. In contrast, we propose to use a pretrained GAN to match face images. To our knowledge, this work is the first to propose image matching with GANs. Besides matching, some recent works have attempted to use neural networks to aid in causal inference, e.g. by learning more invariant / balanced representations~\cite{johansson2016learning,shalit2017estimating,assaad2020counterfactual,zhao2020deep,shi2020invariant,qidong2020new,parascandolo2020learning,mahajan2020domain,kallus2020deepmatch}, using generative models~\cite{kocaoglu2017causalgan,goyal2019explaining,vowels2020targeted,bica2020estimating,averitt2020counterfactual}, learning causal features~\cite{chalupka2017causal,kinney2020causal}, or calculating adjustments~\cite{shi2019adapting,farajtabar2020balance,neto2020counterfactual}.

\paragraph{Projecting images onto a GAN's latent space} Image projection, sometimes referred to as GAN inversion, involves finding a vector in the GAN's latent space that can generate a desired image.
There are broadly two approaches to do so. The first trains an encoder on synthetic training samples to map from the image domain to the latent space~\cite{zhu2020indomain, zhu2016generative, perarnau2016invertible, bau2020semantic}.
A second, optimization-based~ approach \cite{lipton2017precise, ma2018invertibility, abdal2019image2stylegan} uses gradient descent to find a code in the latent space which best generates the image. We build on the second approach, which has been shown to be more stable and generalizable~\cite{karras2019analyzing,abdal2019image2stylegan}.

\paragraph{Analyzing bias in computer vision models}
A long line of work has analyzed bias in computer vision~\cite{barron1994performance,bowyer1998empirical,fei2004learning,brandao2019age} and face analysis~\cite{buolamwini2018gender,klare2012face,lu2019experimental,raji2019actionable,drozdowski2020demographic,kortylewski2018empirically,kortylewski2019analyzing,merler2019diversity,phillips2003face,phillips1998feret,phillips2018face,cavazos2019accuracy,Krishnapriya2019,Howard2019TheEO,el2016face,grother2018ongoing1,grother2019ongoing3}, often focusing on discrepancies in performance (e.g. error rates) across protected attributes (e.g. gender).
The datasets used in these works often contain many sources of possible confounding: combinations of attributes are disproportionately represented and/or correlated~\cite{wang2019balanced,ponce2006dataset,torralba2011unbiased}.

Recent approaches to mitigating dataset bias include collecting more comprehensive samples~\cite{merler2019diversity}, synthesizing images to compensate for distribution gaps~\cite{kortylewski2019analyzing}, weighting examples~\cite{li2019repair}, and explicit annotation~\cite{karkkainen2019fairface}. One study~\cite{muthukumar2018understanding} analyzes images that are manually modified in photoshop to change only the skin color of a face -- however, such an approach does not scale well and may not work for attributes such as gender. Another study~\cite{balakrishnan2020causal} uses a GAN to manipulate the images along only one attribute. This approach is limited in its ability to benchmark facial recognition, since the manipulations can dramatically change identity.
Instead, our matching approach preserves all image properties besides a single attribute used to create groups, improving the ability to mitigate confounding for data gathered `in the wild'. 





\section{Methods}
Our approach consists of two key components: projecting images onto StyleGAN2's latent ``style'' space, and matching images by measuring distances in that style space. At first glance, it may seem odd to use a generative model like a GAN to measure distance. We use a GAN's latent space because recent results have shown that GANs capture a rich representation of images (as evidenced by their image quality), and that their latent spaces naturally disentangle semantic attributes without any supervision. 

\subsection{Projecting images onto StyleGAN2's style space}
\label{sec:projection}
We build on the approach taken in a recent work ~\cite{karras2019analyzing} which optimizes the latent code to minimize the perceptual distance (measured by VGG16 perceptual distance~\cite{johnson2016perceptual,simonyan2014very}) between the original image and the projected image.
The optimization can be performed in StyleGAN2's original restricted style space $\textbf{z} \in \R^{512}$ or in an expanded space formed by concatenating the style spaces of each decoder level, $Z_E \in \R^{18 \times 512}$. 

\begin{figure*}[t!]
    \centering
    \hspace{10pt}\includegraphics[width=0.14\textwidth]{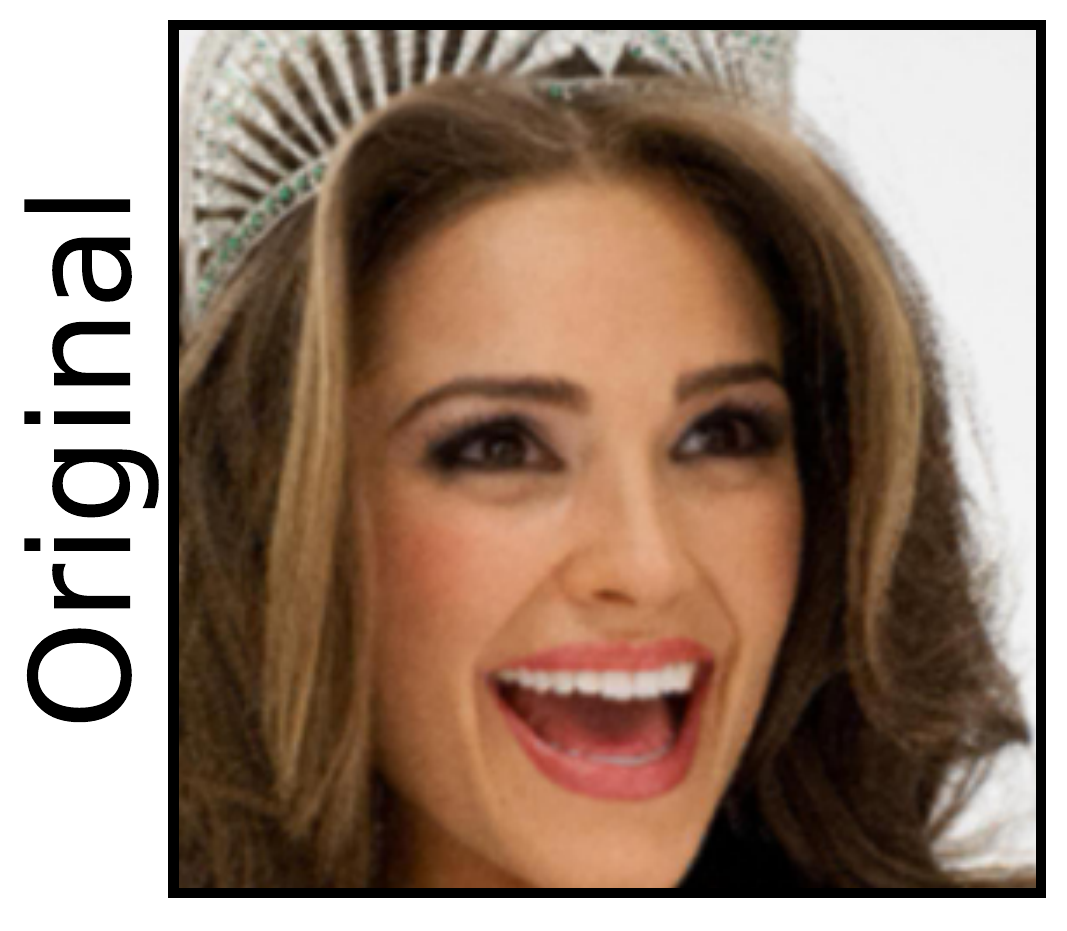}\\
    \raisebox{1.2in}{\rotatebox[origin=t]{90}{\fontsize{0.33cm}{0.33cm}\selectfont(Ours)}}   
    \includegraphics[width=0.9\textwidth]{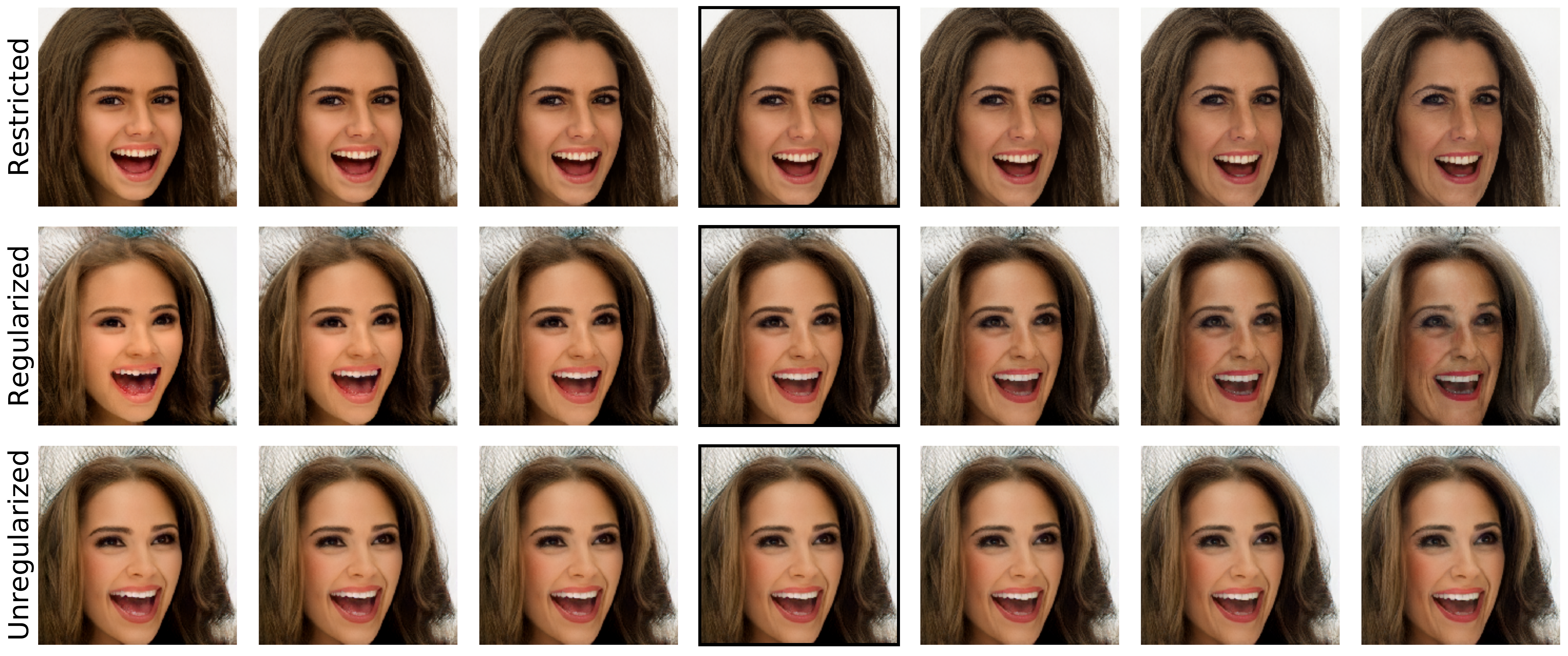} \\
    \vspace{-3pt}
    \hspace{15pt}$ \xleftarrow{\hspace*{4cm}} \hspace{-3pt} \xrightarrow{\hspace*{4cm}} $\\ \vspace{-3pt}
    \hspace{15pt}Manipulating age\\\vspace{3pt}
    \caption{\textbf{Regularized projections achieve good visual reconstruction and sensible semantic manipulation}, when traversing the latent direction associated with age (middle row). The restricted projection (top row) achieves poor visual reconstruction but good semantic manipulation. The unregularized projection in the expanded latent space (bottom row) achieves good visual reconstruction but poor semantic manipulation. Proper regularization ($\lambda = 0.1$, middle row) achieves the best of both worlds. Restricted, regularized, and unregularized latent spaces correspond to those in \cref{fig:projection_stats}.}
    \label{fig:manipulations}
\end{figure*}

\cref{fig:manipulations} shows how reconstruction quality differs in each space for one example face. The middle column shows the original image and its reconstruction in the restricted (top row), and unregularized, expanded (bottom row) spaces. Each row shows the same face manipulated by age using a linear model in the latent space. We trained the linear model using the approach proposed in a previous study~\cite{balakrishnan2020causal} (see \cref{fig:manipulations_skin_color} for another example manipulating skin color). Effective image manipulation with a simple linear model implies that the latent space has good semantic structure, and that distance measurements in the space will be meaningful. Reconstructions using the restricted style space (top row) are poor in terms of perceptual and identity similarity to the original image. Reconstructions using the unregularized, expanded space (bottom row) agree closely with the input image, but semantic manipulation is not possible. A balance between the two extremes is needed.

To explore this tradeoff, we cast the projection task as an optimization problem with the following objective:

\begin{align}
Z_E^* = \argmin_{Z_E} D(G(Z_E), x) + \lambda \sum_j ||Z_{E,j} - \bar{Z}_E||_2^2, 
\end{align}

\noindent where $D(\cdot, \cdot)$ is the perceptual distance measure, $G$ is the GAN generator, $x$ is the image, $||Z_{E,j} - \bar{Z}_E||_2^2$ is a regularizer, and $\lambda$ is a positive scalar. The regularizer penalizes the deviation of each row of the expanded style matrix $Z_E$ to $\bar{Z}_E$, the mean of its rows, effectively pushing the expanded style vectors towards the restricted style space.\footnote{While we use this term in an optimization-based projection approach, it could also be used in an encoding-based projection approach.} Optimization is performed with gradient descent.

\begin{figure}[t!]
    \centering
    \includegraphics[width=\columnwidth]{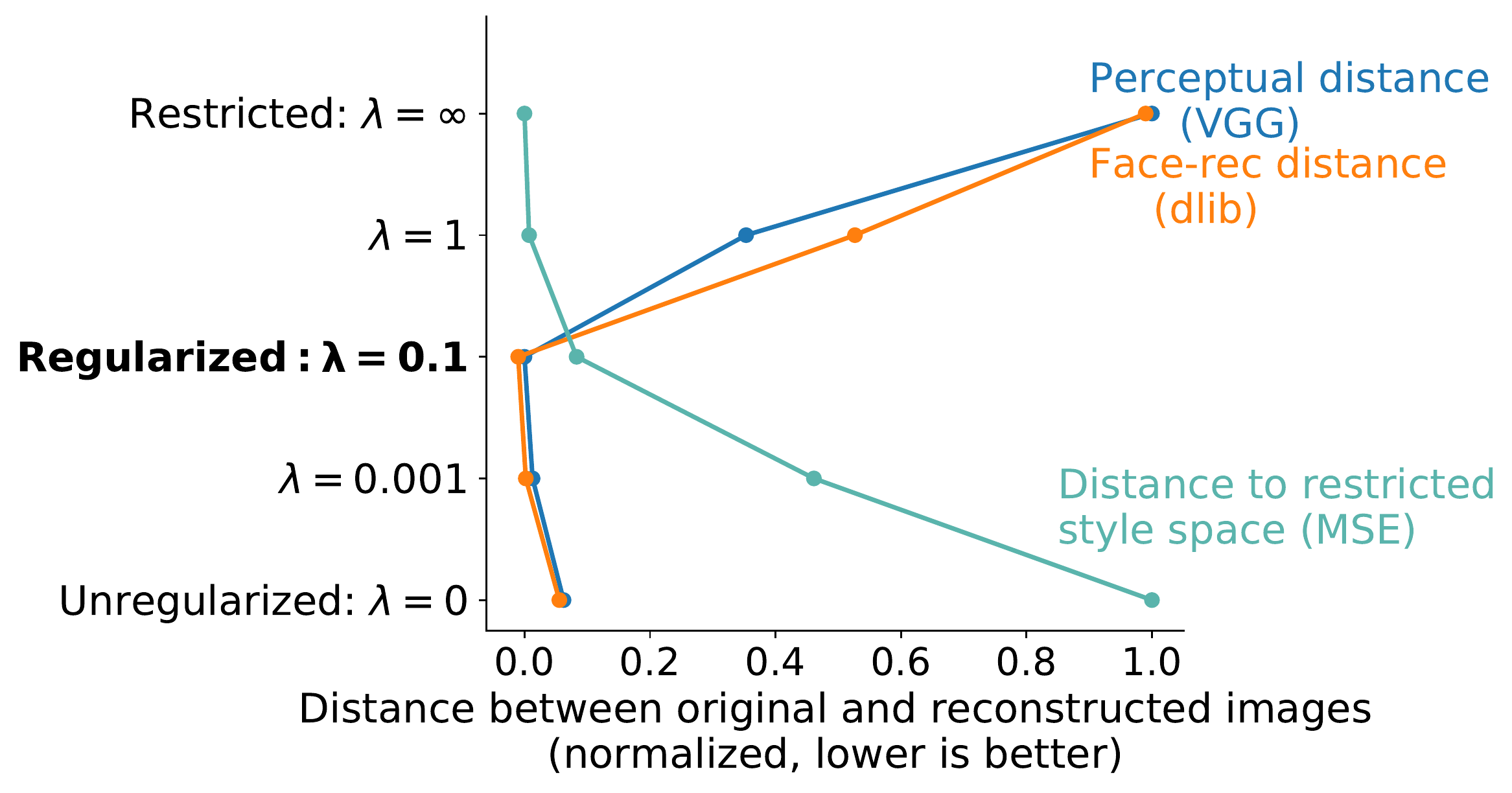}
    \caption{\textbf{Regularization improves projection quality}. Setting $\lambda = 0.1$ yields a modest improvement in both the perceptual distance between the original/reconstructed images (blue curve) and the facial identity distance (orange curve). Increasing regularization brings the latent projections closer to the restricted style space (green curve), as measured by the regularized term $\sum_j ||Z_{E,j} - \bar{Z}_E||_2^2$. This allows the latent space to better preserve semantic properties (see \cref{fig:manipulations}), which are useful for matching. Each curve is normalized to take minimum value 0 and maximum value 1. These results are averaged over the first 300 images in CelebA-HQ.} 
    \label{fig:projection_stats}
\end{figure}

We quantify this tradeoff by sweeping $\lambda$ in \cref{fig:projection_stats}. Reconstructions using the restricted style space are poor in terms of both perceptual distance (blue curve, measured by VGG perceptual distance) and facial id distance (orange curve, measured by the public dlib facial recognition model ~\cite{dlib09}) between the original and reconstructed images. Unregularized reconstructions $Z_E$ yield projections far from the original restricted domain (green curve), making distance comparisons less reliable for matching. Setting $\lambda = 0.1$ achieves both good reconstruction and semantic structure in the latent space. \cref{fig:manipulations} (middle row) shows a visual result of our approach with $\lambda = 0.1$. The example face is both reconstructed well and manipulated along the age attribute.

\subsection{Matching}
\label{sec:matching}

This section presents two techniques for generating matched samples from an image dataset, assuming that the images have been successfully projected onto the GAN's latent space.

\subsubsection{GAN distance matching}
\label{subsec:matching_criteria_nearest_neighbor}

We assume a dataset (e.g., face images) $x_1, ..., x_n$ and specify a single attribute $a$, sometimes referred to as the ``matching attribute'' or ``treatment variable,'' we would like to analyze for bias. In our experiments, $a$ is gender and we find matches based on the following objective: 

\begin{align}
    \label{eq:matching}
    \text{Match}(x_i, A) = \quad &\underset{j}{\text{argmin}} \overbrace{||Z_{E,i} - Z_{E, j}||_{\text{F}}}^{\text{GAN-space dist}} \\ 
    \label{eq:constraint1}
    \text{subject to } &\quad A_j \neq A_i\\
    \label{eq:constraint2}
    \text{\small \color{black!60!white} (optional) \hspace{3pt}} \text{subject to } &\text{\quad Face-Rec}(x_i) \approx \text{Face-Rec}(x_j)\\
    \label{eq:constraint3}
    \text{\small \color{black!60!white} (optional) \hspace{3pt}}\text{subject to } &\text{\quad} \exists\: \text{Valid-References}(x_i, x_j) 
\end{align}

\noindent where $A$ is a vector containing the binary matching attribute value for each example, $Z_{E,i}$ and $Z_{E, j}$ are latent representations in the expanded latent space for a pair of images $x_i$ and $x_j$. The objective \eqref{eq:matching} ensures that matches are near each other in this space. $||\cdot||_F$ is the Frobenius norm.

The first constraint \eqref{eq:constraint1} requires that each match consists of exactly one observation from each group. If the matching attribute $a$ is not identity-preserving (e.g., skin-color or perceived gender), this implicitly enforces that the match does not have the same identity.

While the objective \eqref{eq:matching} is effective at measuring similarity, subtle identity information can be hard to capture using the GAN latent space alone. To address this, we add a second optional constraint \eqref{eq:constraint2} that enforces that a pair of images are similar with regards to a pretrained facial recognition embedding model, Face-Rec$(\cdot)$ (i.e., their distance is below a pre-specified threshold). In our experiments, we use a CNN from the dlib library~\cite{dlib09} which achieves 99.38\% accuracy on the Labeled Faces in the Wild benchmark~\cite{huang2008labeled}. 

The final constraint \eqref{eq:constraint3} is optional and enforces that each image has a valid reference image, i.e., an extra image of the same identity. This is required for facial recognition benchmarking, for which multiple images of each person are needed. 
Given two images ($x_i, x_j$), two new images ($x'_i, x'_j$) are corresponding valid reference images if they preserve the labeled identity of the original images: $ID(x'_i) = ID(x_i)$ and $ID(x'_j) = ID(x_j )$.\footnote{As part of the definition of a valid reference image, we optionally add that it must satisfy certain ``default attributes'' (in our case, this means they must not be wearing eyeglasses).} When there are many potential reference images satisfying these criteria, we select two ``equally difficult'' reference images. This step handles cases such as when there are near-duplicate photos for one celebrity but not the other. We measure difficulty using VGG perceptual distance~\cite{johnson2016perceptual,simonyan2014very}; specifically, we compute the perceptual distance between each reference image and its corresponding test image and then select reference images that are the closest to being equidistant.\footnote{The choice of VGG perceptual distance here should not induce a bias towards either group, since it is based only on low-level features of ImageNet-1K~\cite{deng2009imagenet} rather than a face dataset. Specifically, distance is measured as the $L_1$-distance between feature vectors extracted by the first four layers of VGG-16.} 

\paragraph{Optimal matching} We select matches sequentially, starting with the match obtaining the smallest GAN-distance (sometimes referred to as ``optimal matching''~\cite{rosenbaum2002overt}).
Once a match is selected, all other images with the same identity of both images in the pair are removed from consideration for further matching.

\subsubsection{Propensity score matching}
\label{subsec:matching_propensity}
A well-studied alternative matching approach is propensity-score matching~\cite{rosenbaum1983central,dehejia2002propensity,abadie2016matching}: matching samples based on their predicted probability for a binary protected attribute (e.g., ``perceived maleness'' in our experiments). Matching examples based on this probability, known as the propensity score, can reduce distribution imbalances caused by variables correlated with the protected attribute. Existing work using propensity scores often assume tabular raw data~\cite{abadie2016matching}, which is not immediately applicable to image datasets. We overcome this issue by again leveraging the GAN latent space.

For each image, we compute its (regularized) expanded latent matrix $Z_E \in \R^{18 \times 512}$ and then project this matrix to a vector $\bar{Z}_E \in \R^{512}$ in the restricted style space, by averaging over the expanded dimension. We then train a logistic regression model to predict the probability of the protected attribute value being positive (i.e., the propensity score) from these restricted style space vectors. We verify that these propensity scores are accurate (a prerequisite for matching), finding that they effectively separate perceived gender groups, achieving 98.1\% accuracy when fitted to the whole dataset and 97.15\% 5-fold cross-validated accuracy (full propensity score distribution given in \cref{fig:propensity_scores}). 

Next, we sequentially label matches using the propensity scores. We loop through the smaller of the two protected attribute groups in random order, find the example in the second group with minimal propensity score distance, and accept the match if this distance is within a fixed threshold (0.1 in our experiments). If the match is accepted, we add both images to a list of matches and remove them from further consideration. Otherwise, we discard the original image. In our propensity-score matching experiments, we did not use the facial recognition distance constraint (\ref{eq:constraint2}) or the valid-references constraint (\ref{eq:constraint3}) from our nearest neighbors matching approach, although they may be added if desired.

\section{Results}
We evaluate our methods on the CelebA-HQ~\cite{karras2017progressive} dataset, which contains 30,000 images of 6,216 unique celebrities (3,433 perceived as female and 2,783 perceived as male).
We use the public, pretrained StyleGAN2 model trained on the Flickr-Faces-HQ dataset~\cite{karras2019style}. We use the public face recognition CNN model from the dlib library~\cite{dlib09}. 

We first show that our projection method preserves identity through human evaluation experiments. Next, we demonstrate that our matching strategies recover close matches in CelebA-HQ, despite not having direct access to various attribute labels. Finally, we show how matched samples may be used for tasks like face identification.

\subsection{Projection evaluation} 
We first test how well our projection approach preserves identity using human annotations. In each annotation trial, we show an annotator a pair of images (A and B), both of the same celebrity. Image A is always a real image. Image B is one of two possibilities: a different real image of the same celebrity depicted in image A, or StyleGAN2's reconstruction of another real image of the celebrity. We pose the question ``Does the test photo contain the same person as the real photo or is it a celebrity look-alike?'' (see user interface in \cref{fig:annot_proj}). We purposefully choose this wording so that annotators have a high bar for judging the GAN reconstructions -- a reconstruction must not only be similar to the original, it must be sufficiently similar such that it could not be a different person who looks very much like the original.

We select images from CelebA-HQ by first sorting the celebrities by their number of unique photos in the dataset. We then select the 30 top celebrities in each of the following demographic sets: Black female, Black male, White female, White male. For each celebrity, we have 1 real pair and 1 ``fake'' pair (i.e., one of the photos is a GAN reconstruction). Each pair is annotated 3 times, resulting in a total of 720 annotations. Annotators were paid \$0.024 USD per annotation. Across all annotations, inter-rater agreement (i.e., the probability that two annotators agree on the label) is 0.737, suggesting that the annotations for individual image pairs are fairly reliable.

\cref{tab:human_annotations} shows the resulting percentage of pairs judged by human annotators to be the same person. The overall difference between fake pairs and real pairs is very small, well within 95\% Wilson confidence intervals (first row). This indicates that the reconstructions preserve identity very well. Furthermore, rows 2-4 show that people who say they recognize the celebrity ``Well'' can better discern the GAN reconstructions, suggesting that identity may not be preserved well enough to fool a very familiar observer. Interestingly, these familiar observers are much more likely to report that the pair is a real pair, regardless of whether the pair contains a GAN reconstruction.

\begin{table}[t!]
    \centering
    \small

\begin{tabular}{lll}
\toprule
{} &           Fake pairs &           Real pairs \\
\midrule
All                   &  66.7 (64.14, 69.10) &  68.3 (65.83, 70.73) \\

\midrule
Well            &  86.4 (82.17, 89.78) &  90.5 (86.58, 93.42) \\
Moderately well &  68.7 (62.75, 74.01) &  65.8 (60.02, 71.06) \\
Not at all            &  49.0 (41.93, 56.07) &  46.9 (39.94, 54.06) \\
\bottomrule\\
\end{tabular}

    \caption{\textbf{Percentage of pairs judged by human annotators to be the same person.} Real pairs contain two real images of a celebrity whereas fake pairs contain one real image and one GAN reconstruction. Parentheticals give 95\% Wilson confidence intervals. \textbf{Top row}: The overall difference between percentages for fake pairs and real pairs is extremely small, well within 95\% Wilson confidence intervals. \textbf{Bottom 3 rows}: Annotators who report that they recognize a celebrity `Well' can better discern real pairs from fake pairs than annotators who say they recognize the celebrity `Moderately well' or `Not at all'.}
    \label{tab:human_annotations}
\end{table}

\cref{fig:annot_breakdowns} gives breakdowns for the annotations along with additional annotation results which show that (i) discrepancies between demographic groups are small and (ii) annotators accurately understand the supplied instructions. \cref{fig:annot_human_vs_facerec} suggests that, on average, humans and facial recognition systems struggle with identifying the same celebrities.

\subsection{GAN-distance measure comparisons}
\label{subsec:distance_comparisons}

We first present quantitative analyses in \cref{tab:distance_comparisons} of the two key components of our distance measure: GAN latent space distance (Eq.~\ref{eq:matching}) and face-recognition embedding distance (Eq.~\ref{eq:constraint2}). Please see \cref{sec:qualitative_matching_results} for qualitative results. For a given image, we retrieve the 10 closest matches in CelebA-HQ based on a particular distance, and measure how well certain attributes of an image are preserved. We approximate most of the attributes in \cref{tab:distance_comparisons} with an algorithm rather than with human annotators. We calculate pose attributes (yaw, pitch, and roll) using a CNN proposed in a recent work~\cite{Ruiz_2018}. We calculate background statistics by segmenting the background with a CNN trained for semantic segmentation~\cite{chaurasia2017linknet} and averaging the pixels corresponding to the background segment. We obtain race from a CNN trained to classify four race categories from a diverse dataset of face images~\cite{karkkainen2019fairface}.

Matches obtained using GAN latent space distance alone (\cref{tab:distance_comparisons}, top row) tend to preserve coarse attributes of an image, such as the background and the pose. Matches obtained using GAN-distance between embeddings from a pre-trained facial-recognition classifier tend to preserve identity and attributes related to identity, such as gender and race (middle row). Finally, we combine both distances by finding the closest matches in GAN space subject to the facial-recognition distance being below a threshold of 0.6 (the recommended threshold for the model when classifying whether two faces have the same identity). The combined distance best preserves both global attributes of the image as well as identity-related attributes (bottom row).

\begin{figure*}
    \centering
    \includegraphics[width=\textwidth]{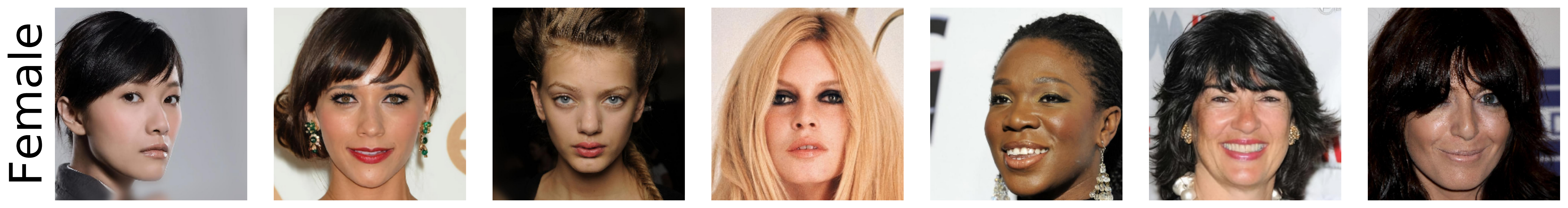}
    \includegraphics[width=\textwidth]{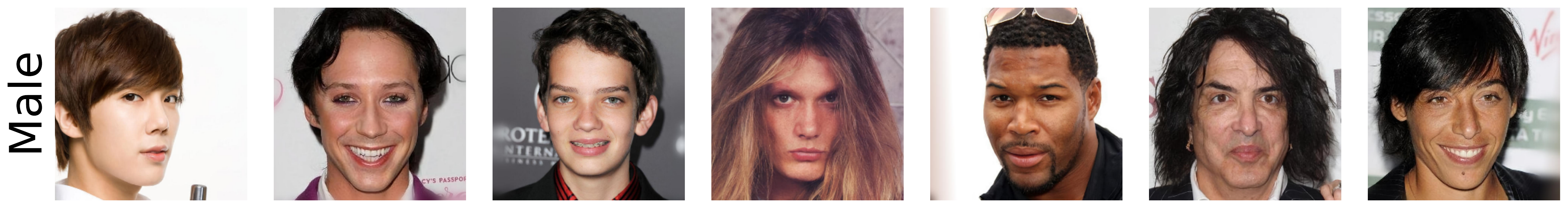}
    \vspace{-10pt}
    \caption{\textbf{Example matches across perceived gender attribute using nearest neighbors matching.}
    Many attributes, such as skin color, hair length, pose, and background texture are preserved. Many more matches shown in \cref{fig:qualitative_gender_matches_full}. Note that the perceived gender may not correspond to a celebrity's self-identified gender.}
    \label{fig:matches_gender}
\end{figure*}

\begin{table*}[h!]
    \centering
    \small
    \begin{tabular}{lllll|lll}
        \toprule
        {Distance measure} & Background Mean &             Yaw &          Pitch &           Roll &        ID (top1) &           Gender &             Race \\
        \midrule
        GAN         &  33.2 $\pm$ 6.0 &   6.6 $\pm$ 2.0 &  5.6 $\pm$ 2.0 &  1.7 $\pm$ 1.0 &  80.4 $\pm$ 20.0 &  16.0 $\pm$ 11.0 &  36.5 $\pm$ 15.0 \\
        Facial recognition &  40.3 $\pm$ 9.0 &  13.8 $\pm$ 4.0 &  6.9 $\pm$ 2.0 &  2.3 $\pm$ 1.0 &   7.8 $\pm$ 13.0 &    1.0 $\pm$ 3.0 &    8.2 $\pm$ 9.0 \\
        Combined        &  38.9 $\pm$ 8.0 &  10.2 $\pm$ 3.0 &  6.5 $\pm$ 2.0 &  2.0 $\pm$ 1.0 &  36.0 $\pm$ 24.0 &    0.9 $\pm$ 3.0 &  14.8 $\pm$ 11.0 \\
        \bottomrule\\
    \end{tabular}

    \caption{\textbf{Errors of attributes between matched images using different distance measures}. Values are errors between an image and its 10 nearest matches in CelebA-HQ based on each distance measure (see \cref{subsec:distance_comparisons}). The values left of the vertical line are mean absolute errors, and the values to the right are percentages. \textbf{Top row}: GAN latent space distance does a good job preserving attributes related to an image's style (i.e., background, face pose). \textbf{Middle row}: In contrast, distance based on a face recognition embedding better preserves identity and related attributes like gender and race. \textbf{Bottom row}: Combining both distances via \cref{eq:matching} and \cref{eq:constraint2} strikes a balance. Intervals are standard errors of the mean.} 
    \label{tab:distance_comparisons}
\end{table*}

\subsection{GAN-distance matching}
\label{sec:nn-matching-results}

Our matching approach (\cref{subsec:matching_criteria_nearest_neighbor}) produces a subset of Celeba-HQ consisting of 1000 images across perceived gender (500 for each group, and 2 photos for each celebrity -- we required 2 per identity for the face recognition benchmarking experiment in \cref{sec:benchmarking}). \cref{fig:matches_gender} shows sample matches. The top row shows faces perceived as female, and the bottom row shows their corresponding matches (see many more matches in \cref{fig:qualitative_gender_matches_full}).
Matches accurately preserve attributes of the face (e.g., skin color, hair length), pose (e.g., yaw, pitch), and background (e.g., color, texture).

\begin{figure*}[h!]
    \centering
    \includegraphics[width=0.9\textwidth]{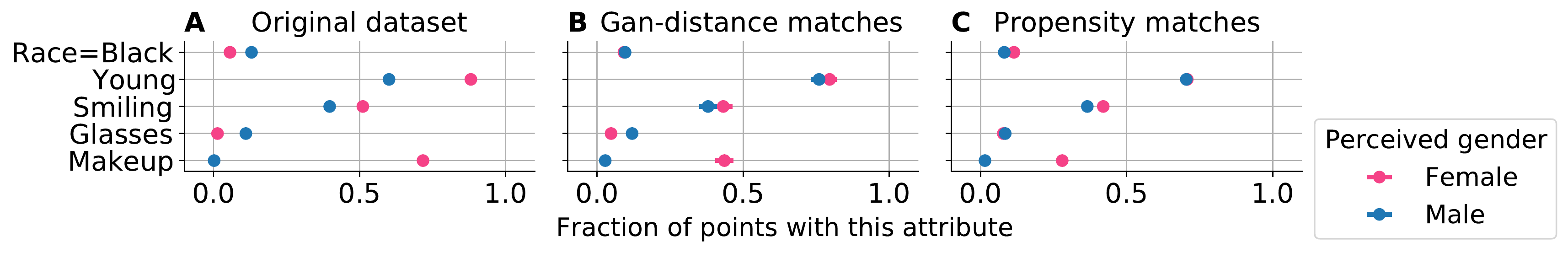}
    \caption{\textbf{After matching, distributions of key covariates are more similar across subgroups.} This is shown by the fact the gap between the means for different gender subgroups is smaller after matching (B, C) than on the original dataset (A). Error bars are 95\% Wilson confidence intervals (often within the points).}
    \label{fig:covariate_matching_full}
\end{figure*}

\cref{fig:covariate_matching_full} shows the effect of matching on the distribution of (binary) key covariates. \cref{fig:covariate_matching_full}A shows the mean value of different covariates for each group in the original dataset. The means are significantly different between groups, indicating the presence of confounding between perceived gender and attributes such as race, age, and smiling. Gaps between these means can skew a downstream analysis of gender bias. After matching, the gaps between these covariates shrink considerably (\cref{fig:covariate_matching_full}B). The covariate means shift closer together except for \textit{Makeup}, which is likely because there exist very few celebrity in CelebA-HQ who are both perceived as male and wear makeup. The result in \cref{fig:covariate_matching_full} shows that GAN-based matching alone can match many important face attributes without requiring access to attribute labels except for the matching attribute (gender in this case).

Importantly, since this procedure matches each observation one-to-one, it does more than simply match the means of individual covariates. It also matches joint distributions between covariates, as indicated by the intersectional distributions in \cref{fig:covariate_intersectional_matching}A. 

\subsection{Propensity-score matching} 
We extract a total of 1,210 unique images (605 per group) using propensity-score matching. \cref{fig:covariate_matching_full}C shows how means of key covariates also shift closer together after matching on propensity scores. While both GAN-distance matching and propensity-score matching substantially reduce covariate gaps between the groups, propensity matching seems to slightly improve balance for attributes such as \textit{Eyeglasses} and \textit{Makeup}.

\begin{figure}[h]
    \centering
    \includegraphics[width=\columnwidth]{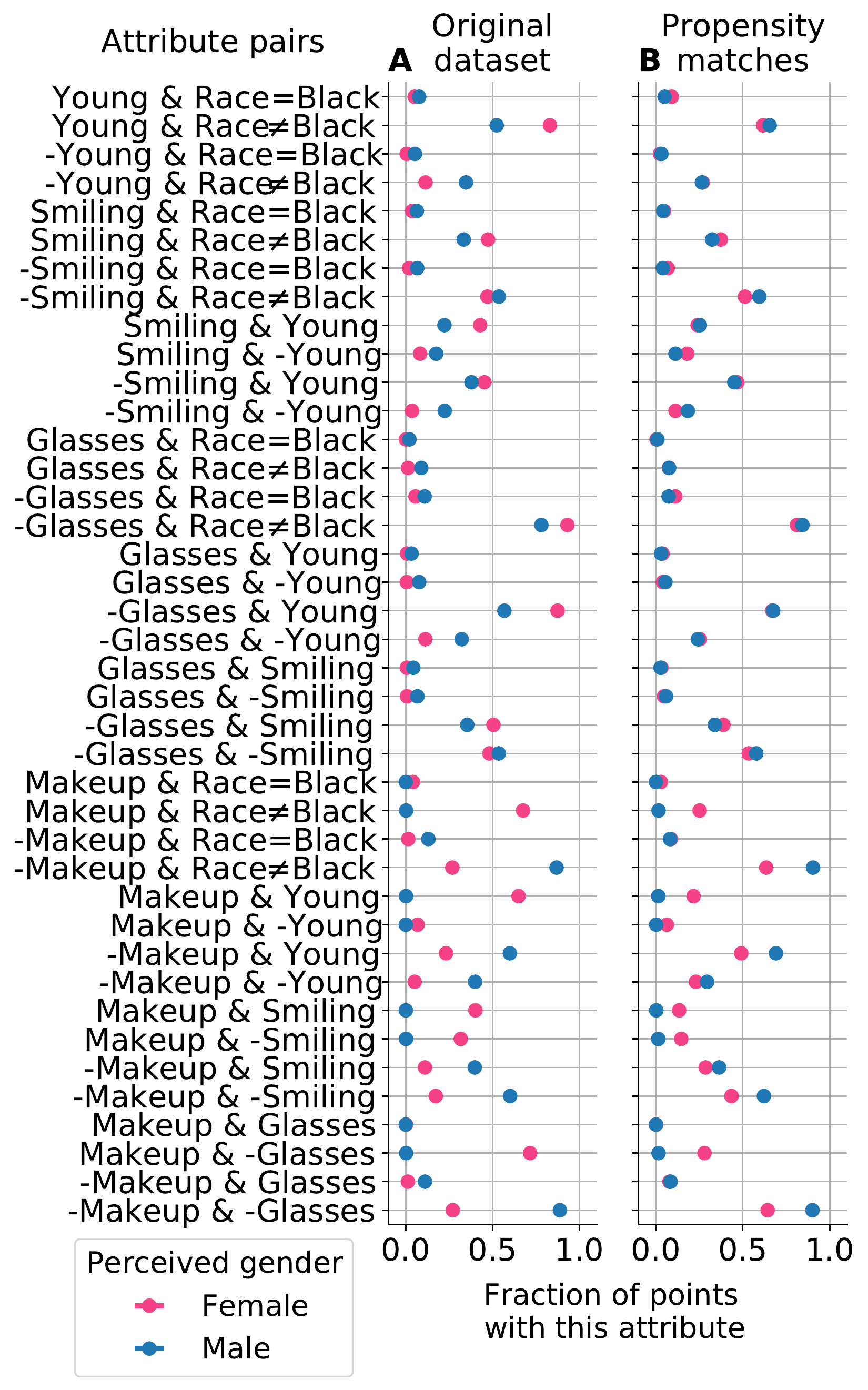}
    \vspace{-10pt}
    \caption{\textbf{After matching, the joint distribution of key covariates becomes more similar between gender groups}. Samples are matched using propensity scores. Error bars are 95\% Wilson confidence intervals (within the points). ``-Young" means that the binary attribute ``Young" is false. \cref{fig:covariate_intersectional_matching} shows similar results for the nearest-neighbor matches.}
    \label{fig:covariate_intersectional_matching_propensity}
\end{figure}

Again, we are interested in not only matching individual distributions of covariates but also their joint distributions. \cref{fig:covariate_intersectional_matching_propensity}B shows the effect of matching on the joint distributions for key covariates. Across the board, propensity score matching reduces the gap between the groups for different covariate combinations, sometimes quite substantially. For example, propensity matching removes the large imbalance for the \textit{Young \& Race $\neq$ Black} category between gender groups. 

\subsection{Benchmarking facial recognition models on matched samples}
\label{sec:benchmarking}
Finally, we show how matching may be used when benchmarking bias in facial recognition systems. Matching can extract a more balanced subset of a large dataset collected in the wild like Celeba-HQ. We use the GAN-distance matched samples from \cref{sec:nn-matching-results} and benchmark the following popular, open-source systems: dlib~\cite{dlib09}, Inception ResNet v1~\cite{szegedy2017inception} trained on VGGFace2~\cite{cao2018vggface2} and CASIA-WebFace~\cite{yi2014learning}\footnote{Facenet models retrieved from \url{https://github.com/davidsandberg/facenet}.}. Each model returns a face embedding vector for a face, and the distance between any two embeddings is a measure of identity similarity. An accurate recognition model should report a small distance between images of the same identity.

\begin{figure}[h!]
    \centering
    \includegraphics[width=\columnwidth]{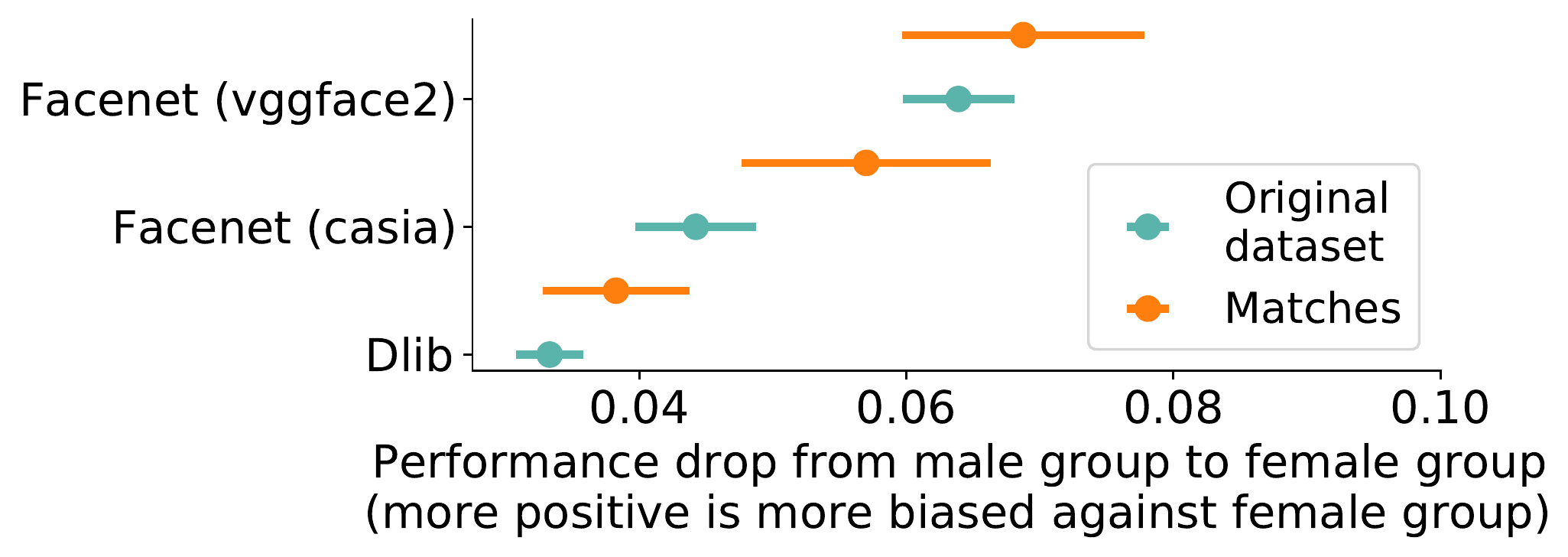}
    \vspace{-10pt}
    \caption{\textbf{All classifiers perform worse for female celebrities}, both before matching (Original dataset) and after GAN-distance matching (Matches). We quantify the performance of a classifier by its mean facial embedding distance for image pairs with the same identity, which should be small. Error bars are standard errors of the mean for the female subgroup.}
    \label{fig:bias_benchmarking}
\end{figure}

For each perceived gender group, we average the distances reported by each recognition model between images of the same identity. \cref{fig:bias_benchmarking} presents the difference between mean recognition distances for females and males (a positive value means females have larger distances on average than males). The original dataset refers to benchmarking with the full CelebA-HQ dataset. Differences are positive both before and after matching, indicating that all models perform worse for females. Furthermore, the magnitude of the differences are greater using the matched samples, suggesting that confounding factors in the full Celeba-HQ data may mask gender bias. This result also shows how the composition of the benchmarking dataset is critical to accurately measure bias.

\section{Discussion and conclusions}


We propose a GAN-based matching method which returns matched samples that are both visually accurate (\cref{fig:matches_gender}) and balanced across attributes (\cref{fig:covariate_matching_full}, \cref{fig:covariate_intersectional_matching_propensity}). Our approach relies on no supervision except for labels of the matched attribute.
We focused on faces in this work because of their prevalence in sensitive applications, and the impressive performance of StyleGAN for this domain. However, our ideas will likely be applicable to any domain where GANs can learn meaningful latent spaces, such as in bioimaging. 

Our regularization-based technique for projecting images onto a GAN latent space improves visual reconstruction over past approaches, as measured using perceptual metrics (\cref{fig:projection_stats}) and through human annotation experiments (\cref{tab:human_annotations}), an important step missing in existing work. Additionally, our projections preserve semantic properties useful for matching (\cref{fig:projection_stats}, \cref{fig:manipulations}) and can be used for downstream tasks such as GAN-based semantic image editing and style transfer, independent of their use for matching.
They can also serve as a disentangled space in which to interpret~\cite{singh2018hierarchical,singh2020transformation} and improve a downstream classifier~\cite{rieger2020interpretations}.

We applied our method to the measurement of algorithmic bias in facial recognition and found that these models have higher error rates on celebrities perceived as female. Unlike previous methods for measuring algorithmic bias, our method mitigates spurious attribute correlations that may bias the measurement, as is the case with many state-of-the-art observational studies. There are many future challenges to improve causal analysis of images, and the study here helps set a course for more rigorous benchmarking of bias on images ``in the wild''. 

\iftoggle{cvprfinal}{
    \section*{Acknowledgements}

The authors would like to thank Luis Goncalves for very useful discussions and comments. Additionaly, we would like to thank De'Aira Bryant, Nashlie Sephus, Wei Xia, Yuanjun Xiong and the rest of the faces team and fairness team at Amazon for thoughtful feedback and discussions.
}

\FloatBarrier
{
    \iftoggle{cvprfinal}{\footnotesize}{\small}

    \iftoggle{cvprfinal}{\bibliographystyle{unsrt}}{\bibliographystyle{ieee_fullname}}
}

{
    \onecolumn
    \appendix
    \etoctocstyle{1}{Appendix}
    \etocdepthtag.toc{mtappendix}
    \etocsettagdepth{mtchapter}{none}
    \etocsettagdepth{mtappendix}{subsection}
    \tableofcontents
    \resetcounters
    \section{Projection results continued}

\begin{figure}[H]
    \centering
    \includegraphics[width=0.14\textwidth]{figs/fig_manipulations_orig.pdf}
    \includegraphics[width=0.9\textwidth]{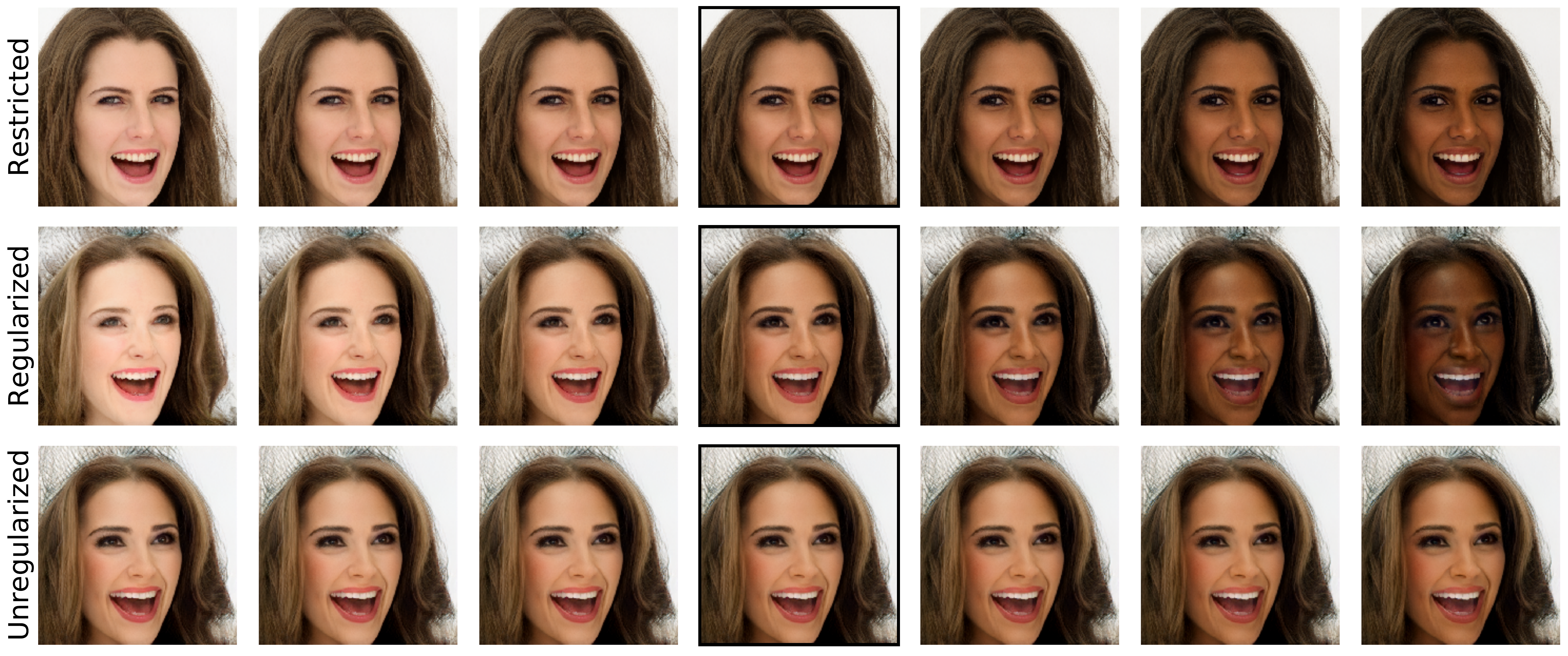}
    \caption{Regularization achieves good visual reconstruction and allows for semantic image manipulation. Middle column shows that the restricted latent space does not qualitatively look like the original image, but the expanded latent space (with and without regularization) both do. In the regularized latent space, traversing the latent direction associated with skin color does change the face skin color, whereas the unregularized version does not.}
    \label{fig:manipulations_skin_color}
\end{figure}

\subsection{Human annotation experiment details}

\label{subsec:annot_gan_details}
\begin{figure}[H]
    \centering
    \includegraphics[width=0.8\textwidth]{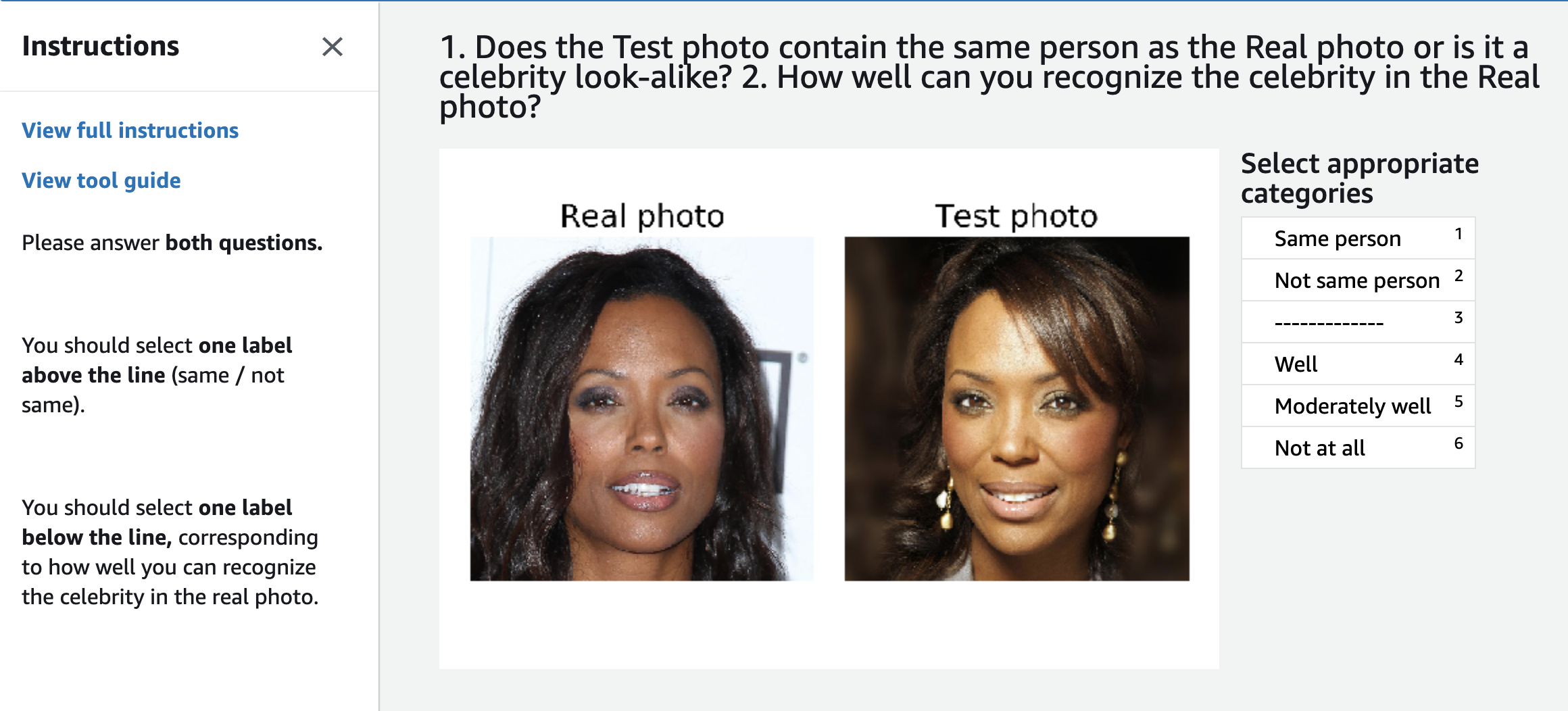}
    \caption{Online interface for benchmarking annotation projections.}
    \label{fig:annot_proj}
\end{figure}

\begin{figure}[H]
    \centering
    \includegraphics[width=0.39\textwidth]{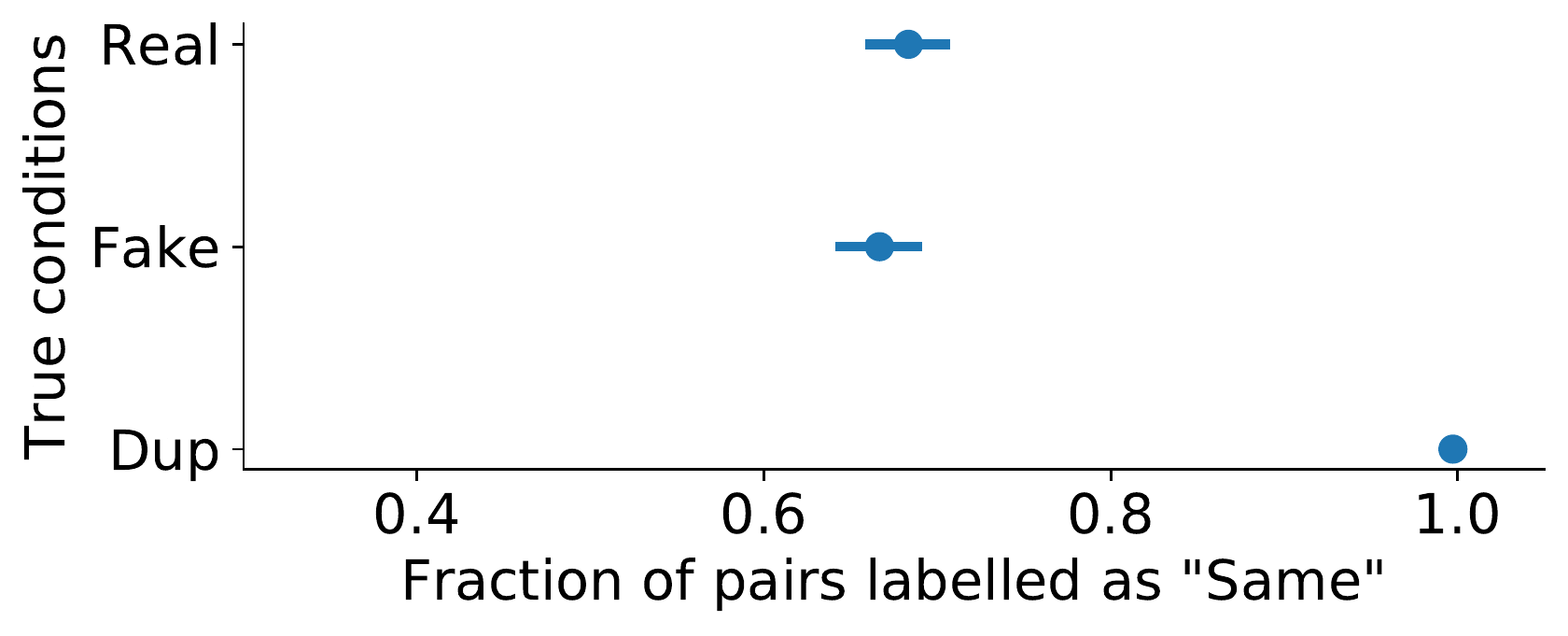}
    \includegraphics[width=0.6\textwidth]{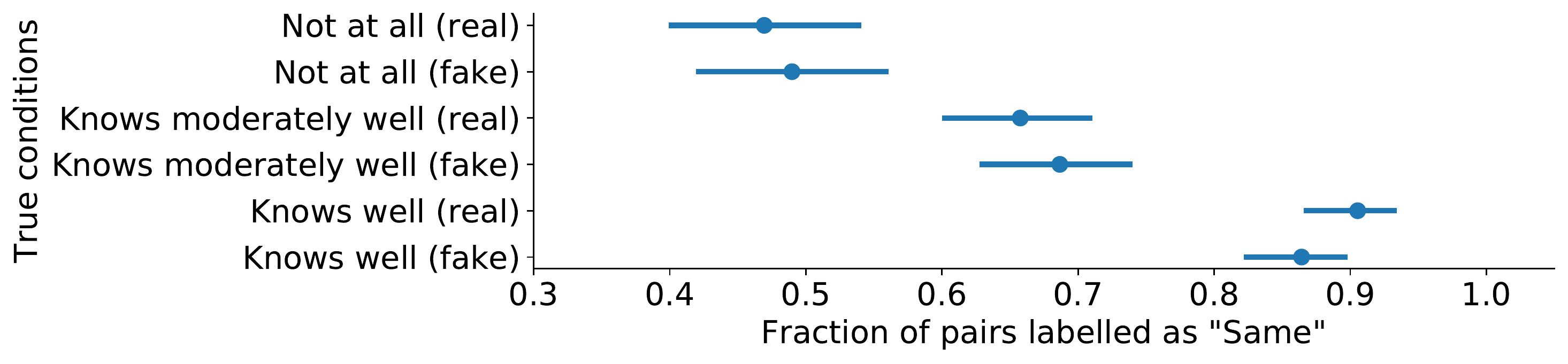}
    \includegraphics[width=0.45\textwidth]{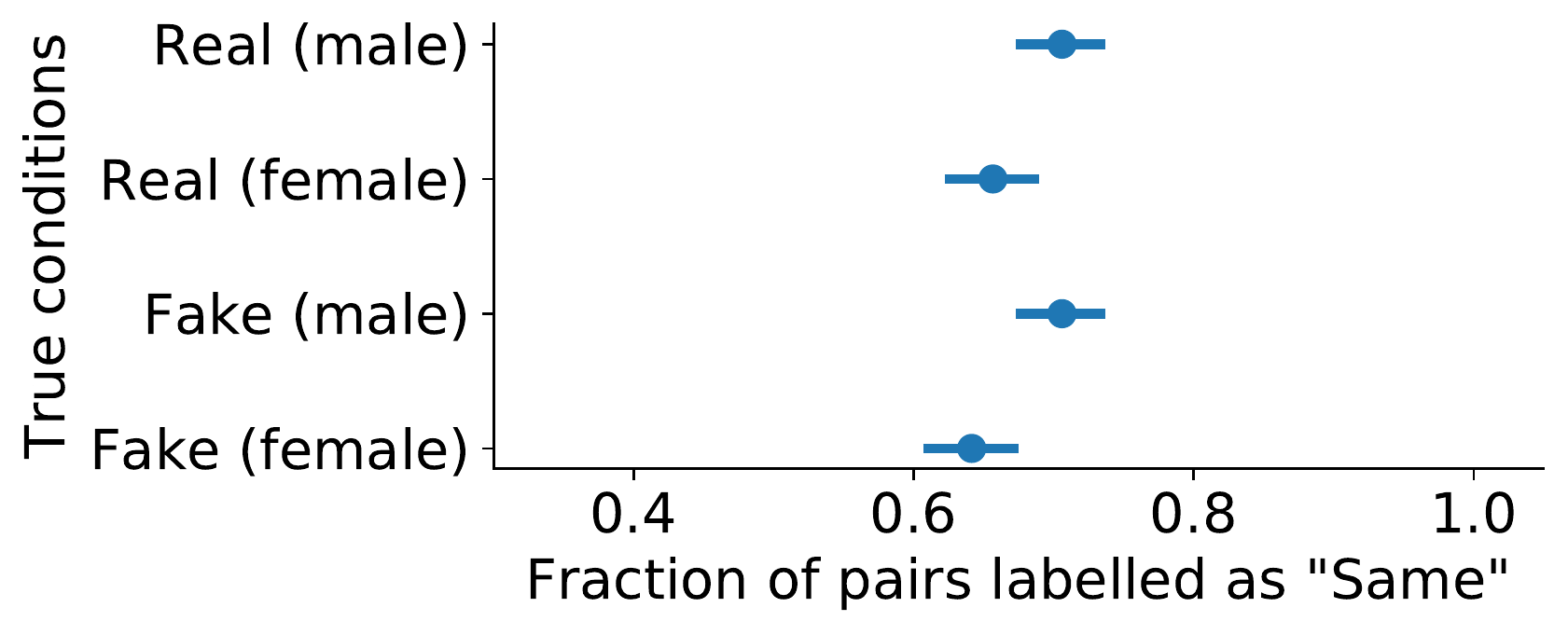}
    \includegraphics[width=0.45\textwidth]{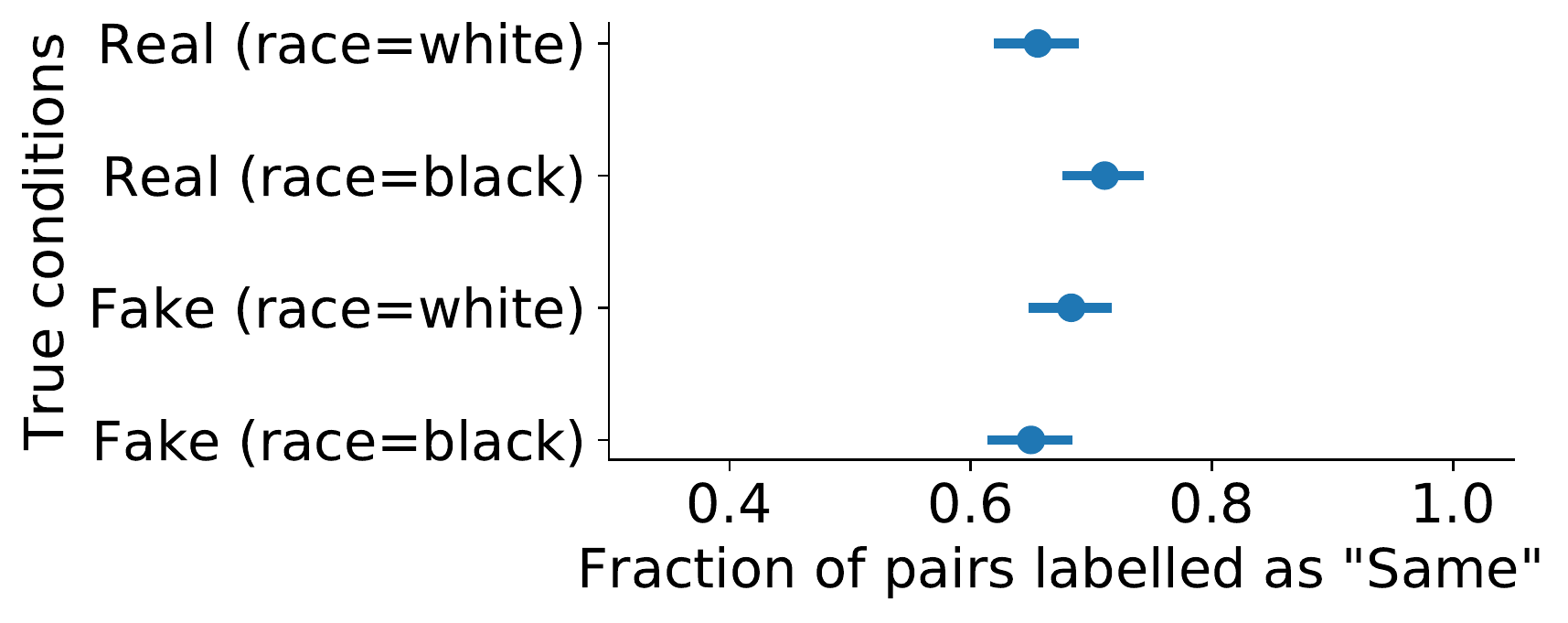}
    \caption{Annotation breakdown. \textbf{Top-left.} Duplicate image (`Dup') shows that people accurately detect when the photos are duplicates with small crops. There is almost no difference between real and fake. People who know celebs better do better. \textbf{Top-right} People who knew the celebrity well were slightly more accurate at identifying the real celebrity photo as opposed to the GAN reconstructed photo. Interestingly, they are also more likely to label the pair  of images as the same (regardless of whether it was GAN-reconstructed or not). \textbf{Bottom-left} Error rates are roughly equal across male/female celebrities, although female celebrities are slightly less likely to be labelled as the ``Same''. \textbf{Bottom-right} Error rates for black celebrities are slightly lower than for white celebrities. All error bars are Wilson confidence intervals.}
    \label{fig:annot_breakdowns}
\end{figure}

\begin{figure}[H]
    \centering
    \includegraphics[width=0.39\textwidth]{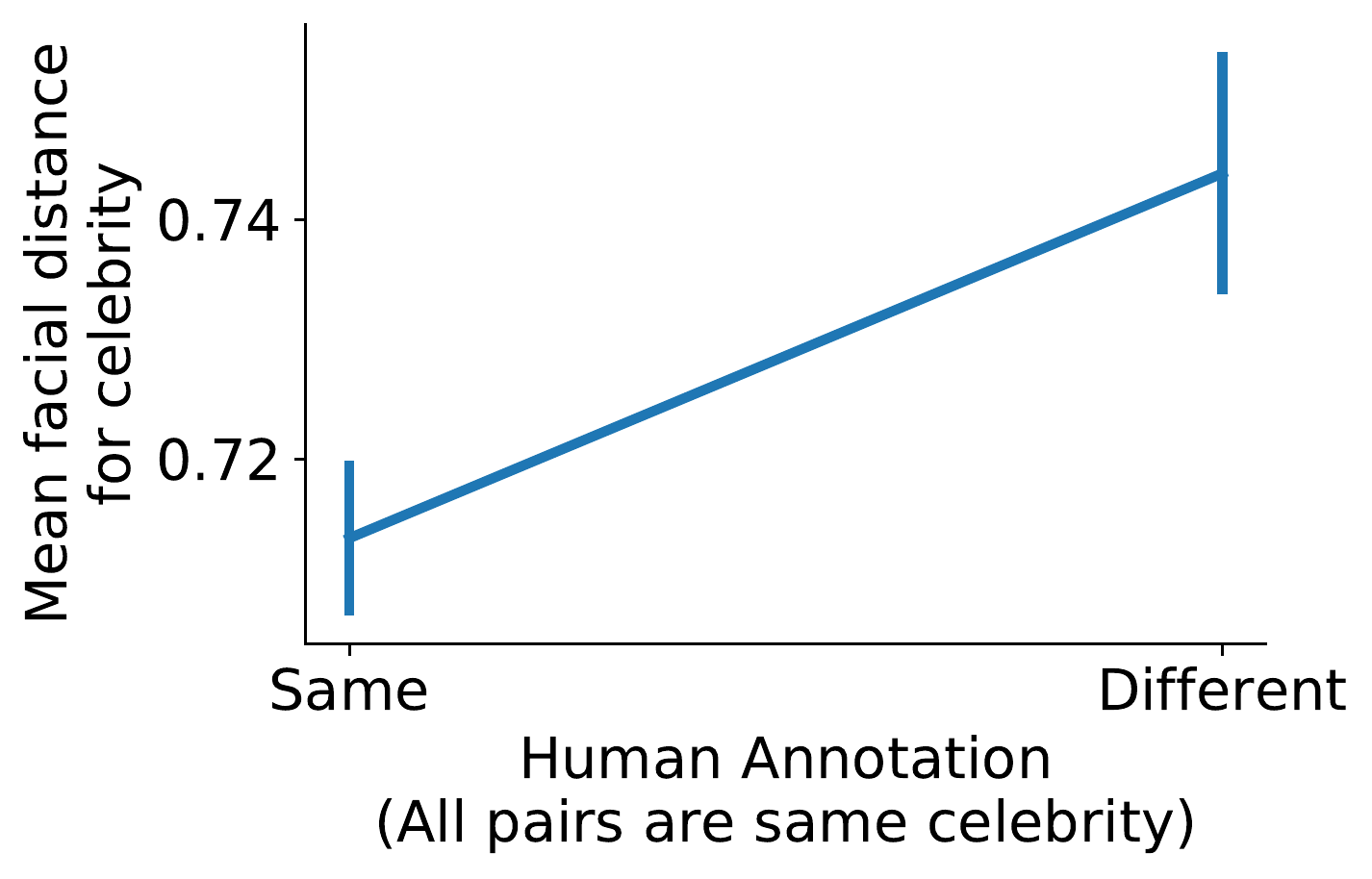}
    \caption{Facial recognition and humans struggle with the same celebrities. Pairs that human annotators annotated as being the `Same' person have a lower mean facial distance, as measured by a facial recognition classifier. Mean facial distance measured by FaceNet trained on VGGFace2.}
    \label{fig:annot_human_vs_facerec}
\end{figure}
    \resetcounters
    \section{Qualitative matching results}
\label{sec:qualitative_matching_results}

\begin{figure}[H]
    \centering
    \includegraphics[width=0.99\textwidth]{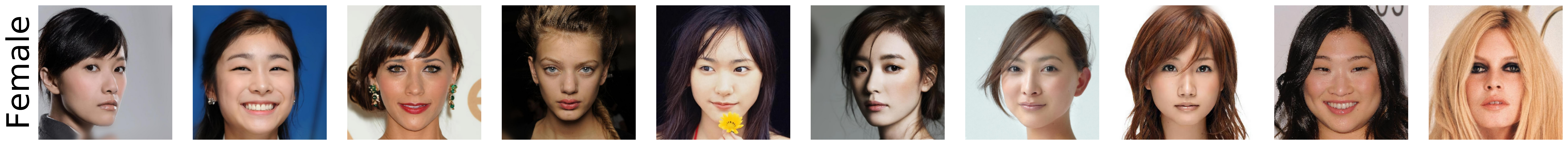}
    \includegraphics[width=0.99\textwidth]{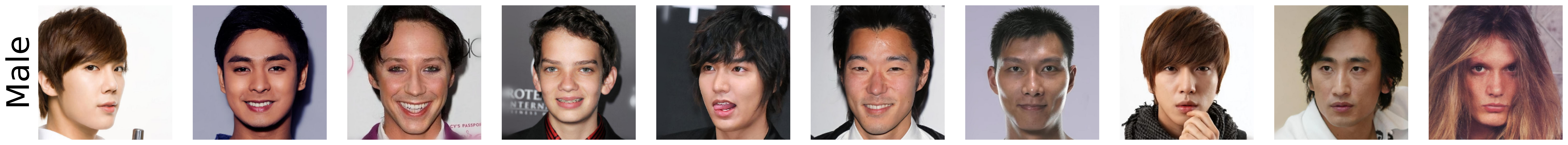}
    \hrule
    \includegraphics[width=0.99\textwidth]{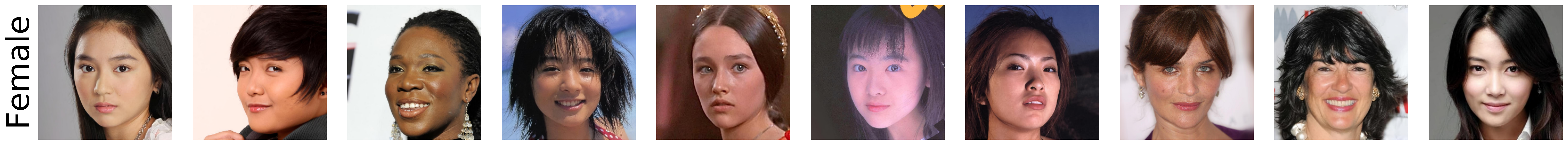}
    \includegraphics[width=0.99\textwidth]{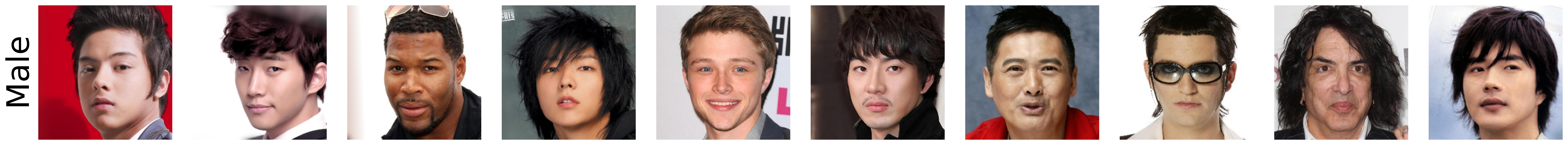}
    \hrule
    \includegraphics[width=0.99\textwidth]{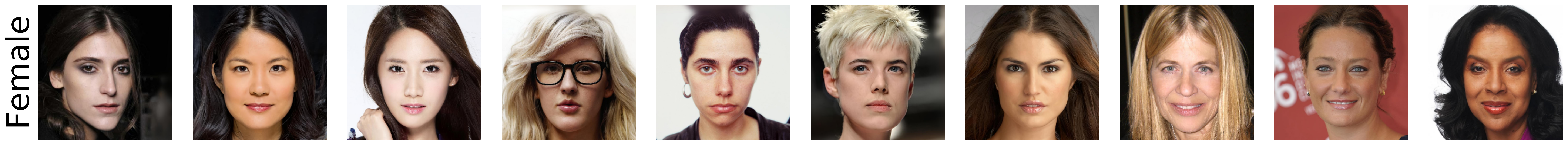}
    \includegraphics[width=0.99\textwidth]{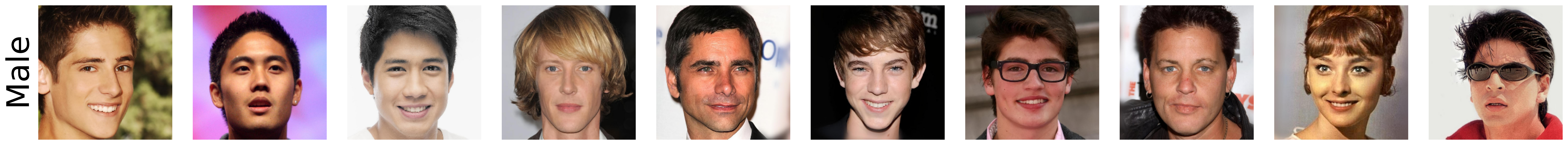}
    \hrule
    \includegraphics[width=0.99\textwidth]{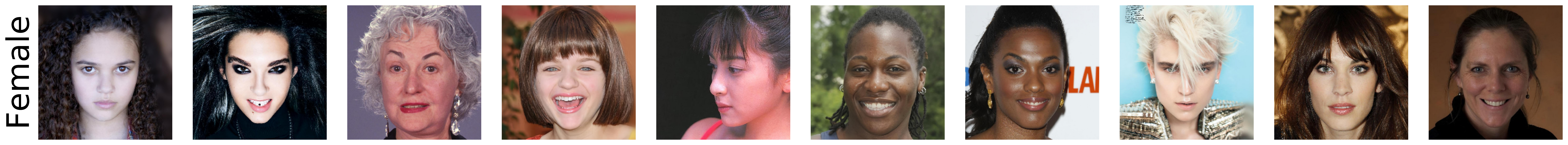}
    \includegraphics[width=0.99\textwidth]{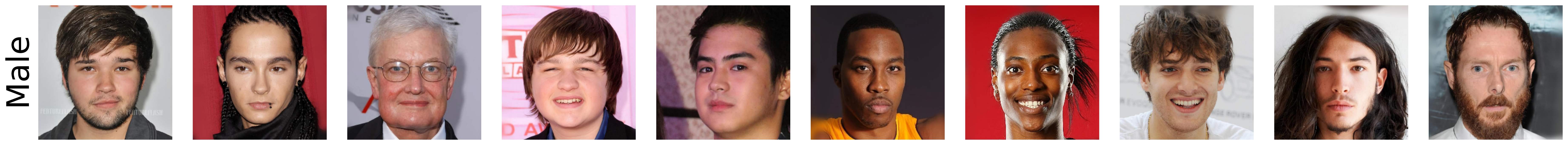}
    \hrule
    \includegraphics[width=0.99\textwidth]{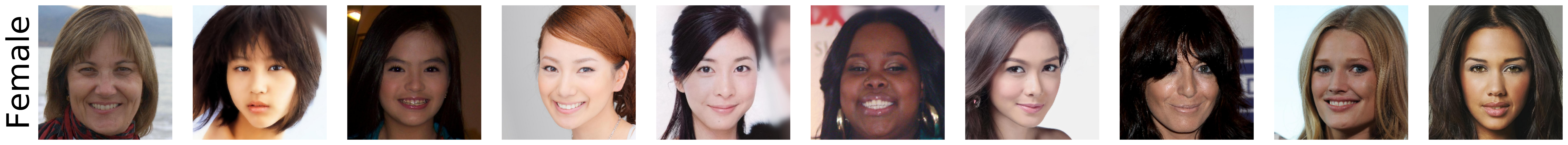}
    \includegraphics[width=0.99\textwidth]{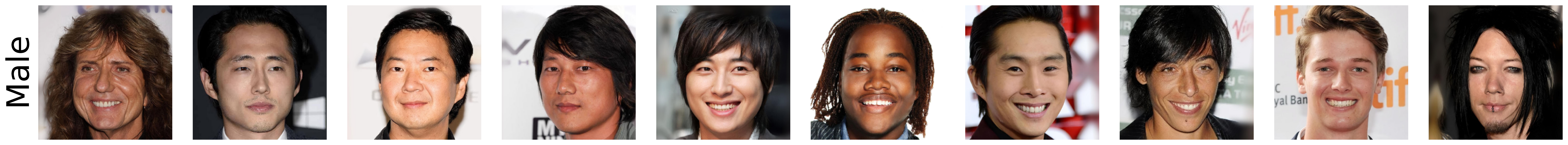}
    \hrule
    \includegraphics[width=0.99\textwidth]{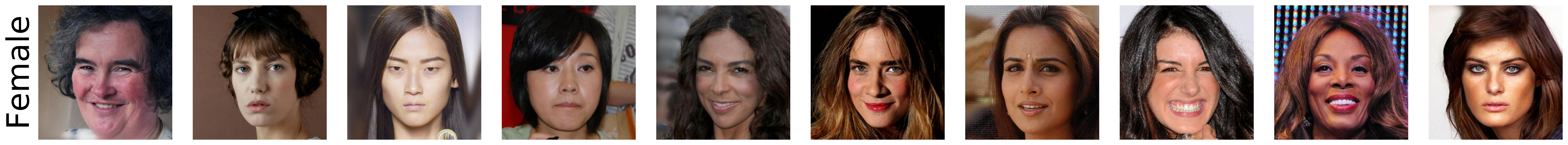}
    \includegraphics[width=0.99\textwidth]{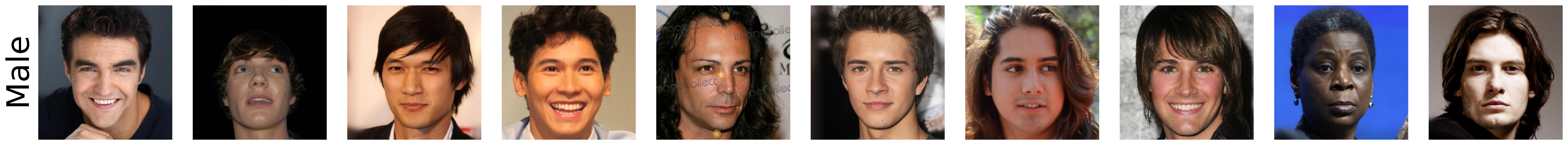}
    \caption{Top matches across perceived gender. Note that the perceived gender may not correspond to the celebrities self-identified gender.}
    \label{fig:qualitative_gender_matches_full}
\end{figure}



\begin{figure}[H]
    \centering
    \includegraphics[width=0.99\textwidth]{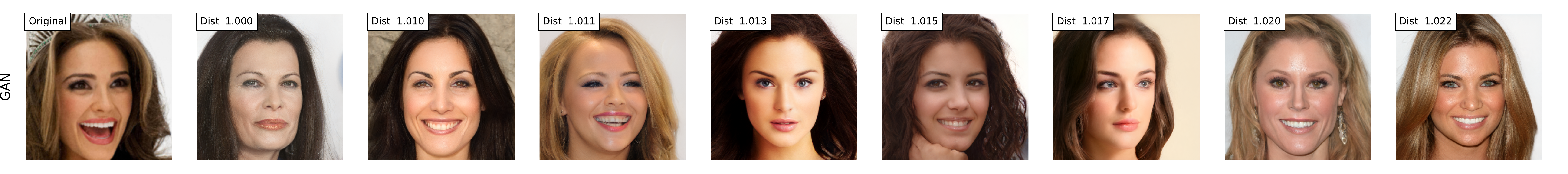}
    \includegraphics[width=0.99\textwidth]{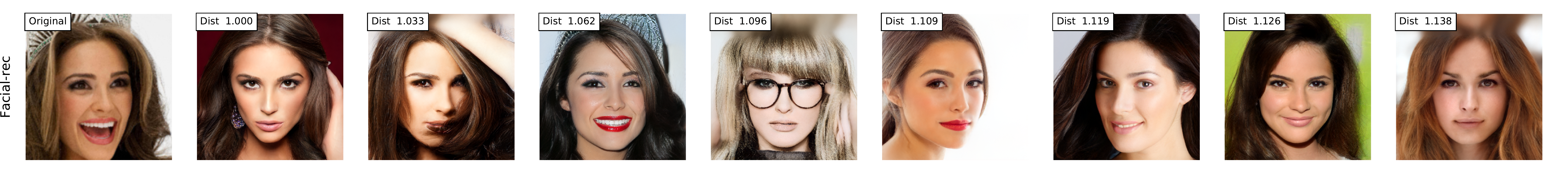}
    \includegraphics[width=0.99\textwidth]{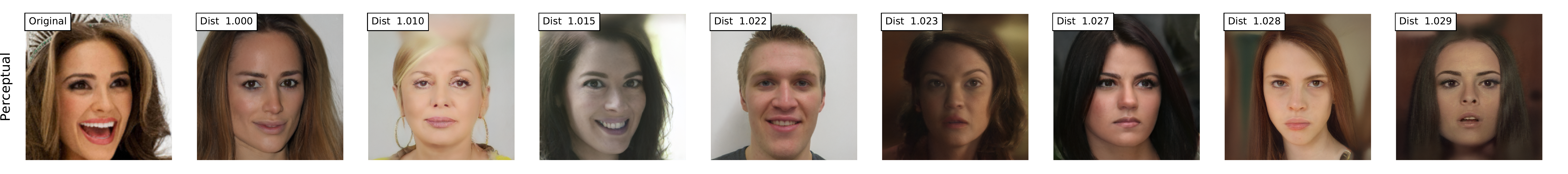}
    \caption{Comparing different distance metrics for matching. Leftmost column shows the original image. Next columns show the top matches (box shows distance from original image, where the distances are normalized by dividing the distance to the closest matches to make distance values comparable across rows.}
    \label{fig:qualitative_distance_comparisons}
\end{figure}

    \resetcounters
    \section{Quantitative matching results continued}
\label{sec:quantitative_matching_results_cont}
\begin{figure}[H]
    \centering
    \includegraphics[width=0.7\textwidth]{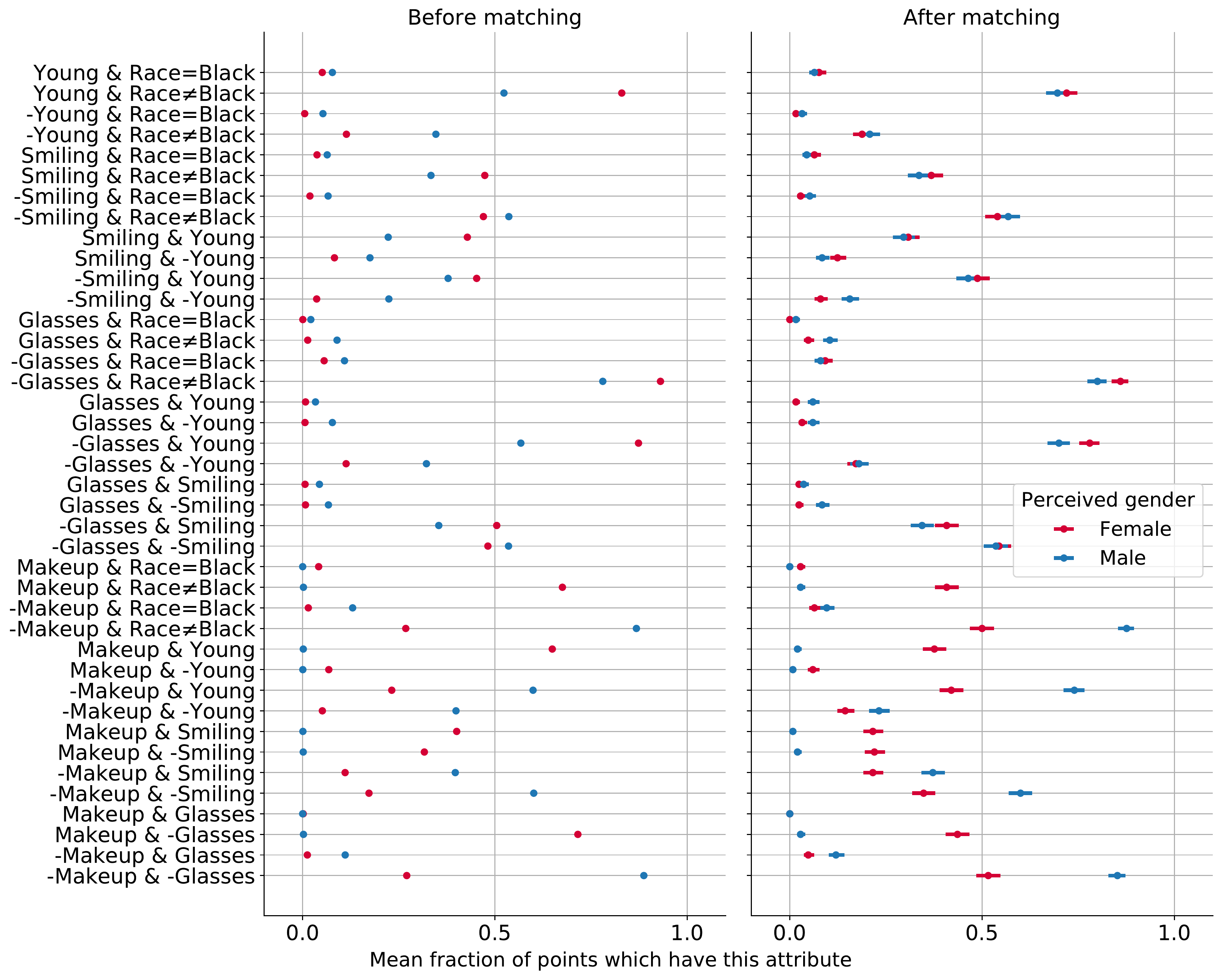}
    \caption{After matching, key covariates (such as the proportion of Black celebrities) are more similar across subgroups. Error bars are 95\% Wilson confidence intervals (often within the points).}
    \label{fig:covariate_intersectional_matching}
\end{figure}

\begin{figure}[H]
    \centering
    \includegraphics[width=0.4\textwidth]{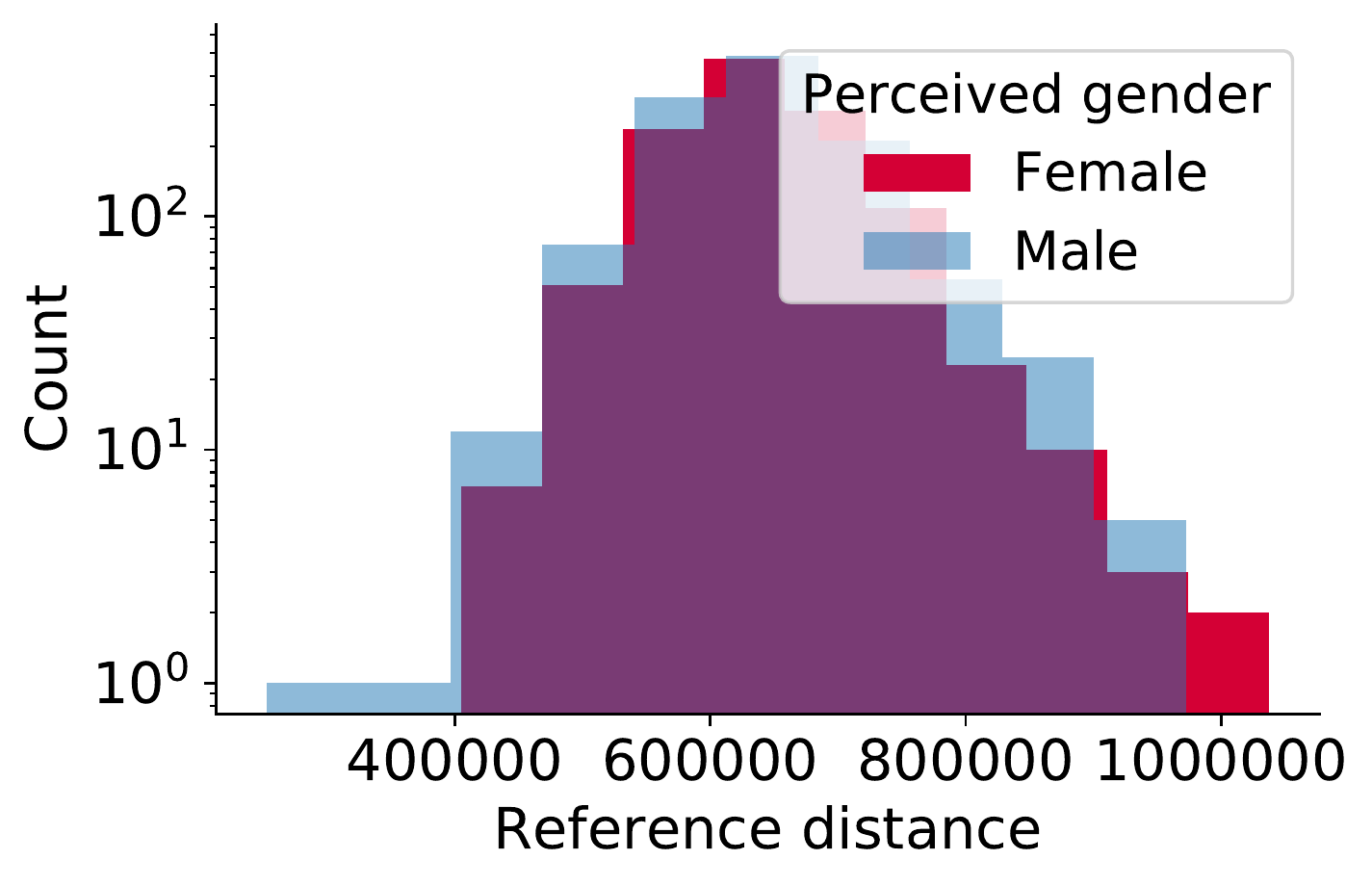}
    \caption{Distribution of reference distances for both groups are similar.}
    \label{fig:ref_dists}
\end{figure}

\begin{figure}[H]
    \centering
    \includegraphics[width=0.4\textwidth]{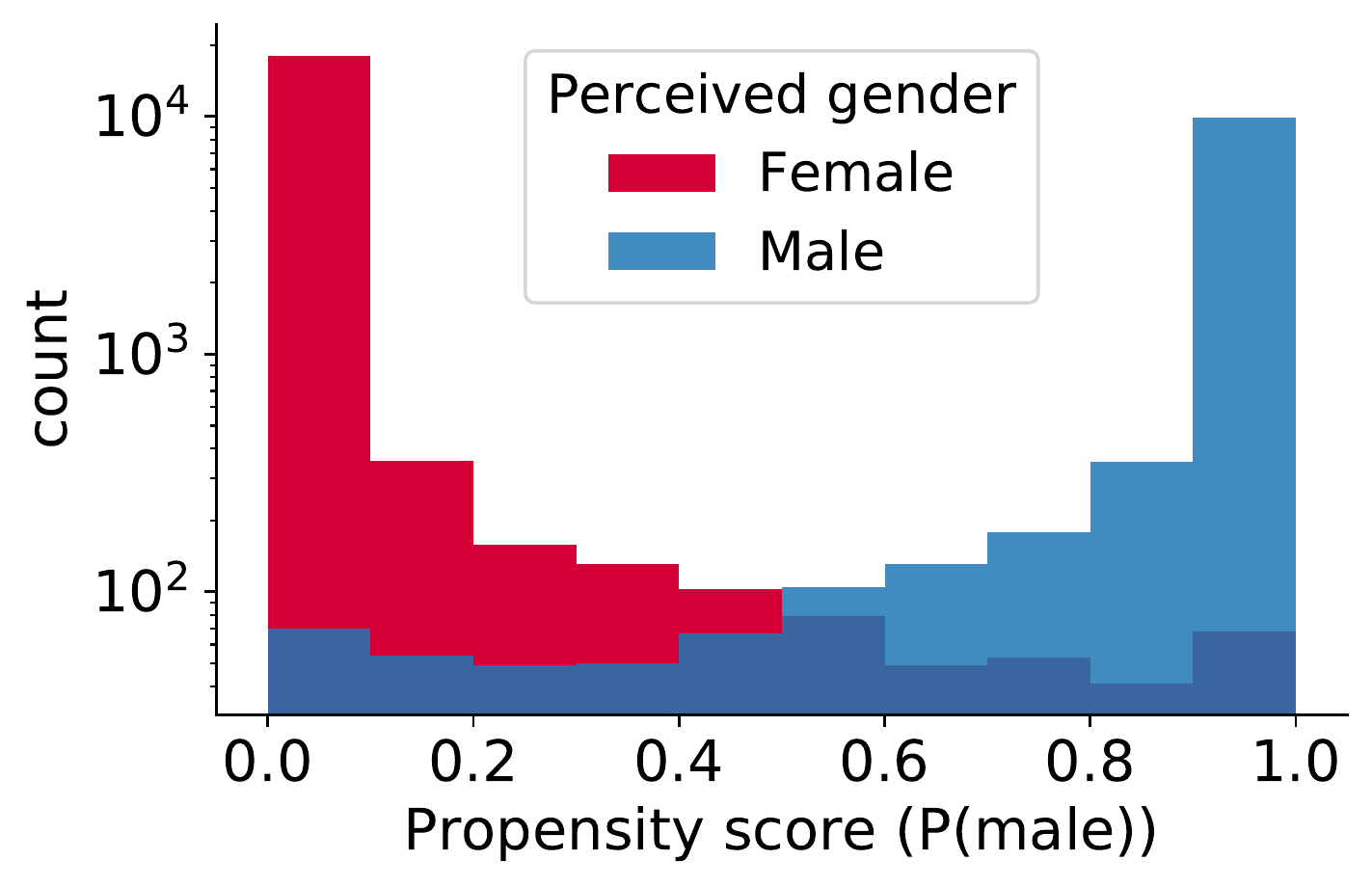}
    \caption{Propensity scores accurately divide the classes.}
    \label{fig:propensity_scores}
\end{figure}
}

\end{document}